\documentclass{article}

\PassOptionsToPackage{numbers,compress}{natbib}
\usepackage[preprint]{neurips_2024}

\usepackage{amsmath}

\usepackage[T1]{fontenc}    
\usepackage{hyperref}       
\usepackage{url}            
\usepackage{booktabs}       
\usepackage{amsfonts}       
\usepackage{nicefrac}       
\usepackage{microtype}      
\usepackage{graphicx}       

\usepackage{float}

\usepackage[most]{tcolorbox} 
\tcbset{fonttitle=\bfseries}

\usepackage{xcolor}
\usepackage{enumitem}
\usepackage{listings}

\newtcolorbox{modelbox}[2][]{%
  enhanced, breakable, drop shadow, arc=2mm,
  left=2mm,right=2mm,top=2mm,bottom=2mm, boxrule=0.8pt,
  title={#1},
  colback=#2!5, colframe=#2!60!black,
  colbacktitle=#2!25, coltitle=black,
  attach boxed title to top left={yshift=-2mm, xshift=2mm},
  boxed title style={arc=1mm, boxrule=0.8pt}
}

\title{On the Limits of Innate Planning in Large Language Models}

\author{%
  Charles Schepanowski \\
  Western University\\
  \texttt{cschepan@uwo.ca} \\
  \And
  Charles Ling \\
  Western University\\
  \texttt{charles.ling@uwo.ca}
}

\begin{document}

\maketitle

\begin{abstract}

Large language models (LLMs) achieve impressive results on many benchmarks, yet their capacity for planning and stateful reasoning remains unclear. We study these abilities directly, without code execution or other tools, using the 8-puzzle: a classic task that requires state tracking and goal-directed planning while allowing precise, step-by-step evaluation. Four models are tested under common prompting conditions (Zero-Shot, Chain-of-Thought, Algorithm-of-Thought) and with tiered corrective feedback. Feedback improves success rates for some model–prompt combinations, but many successful runs are long, computationally expensive, and indirect. We then examine the models with an external move validator that provides only valid moves. Despite this level of assistance, none of the models solve any puzzles in this setting. Qualitative analysis reveals two dominant deficits across all models: (1) brittle internal state representations, leading to frequent invalid moves, and (2) weak heuristic planning, with models entering loops or selecting actions that do not reduce the distance to the goal state. These findings indicate that, in the absence of external tools such as code interpreters, current LLMs have substantial limitations in planning and that further progress may require mechanisms for maintaining explicit state and performing structured search.

\end{abstract}

\section{Introduction}

\begin{figure}[t]
  \centering
  \includegraphics[width=1\linewidth]{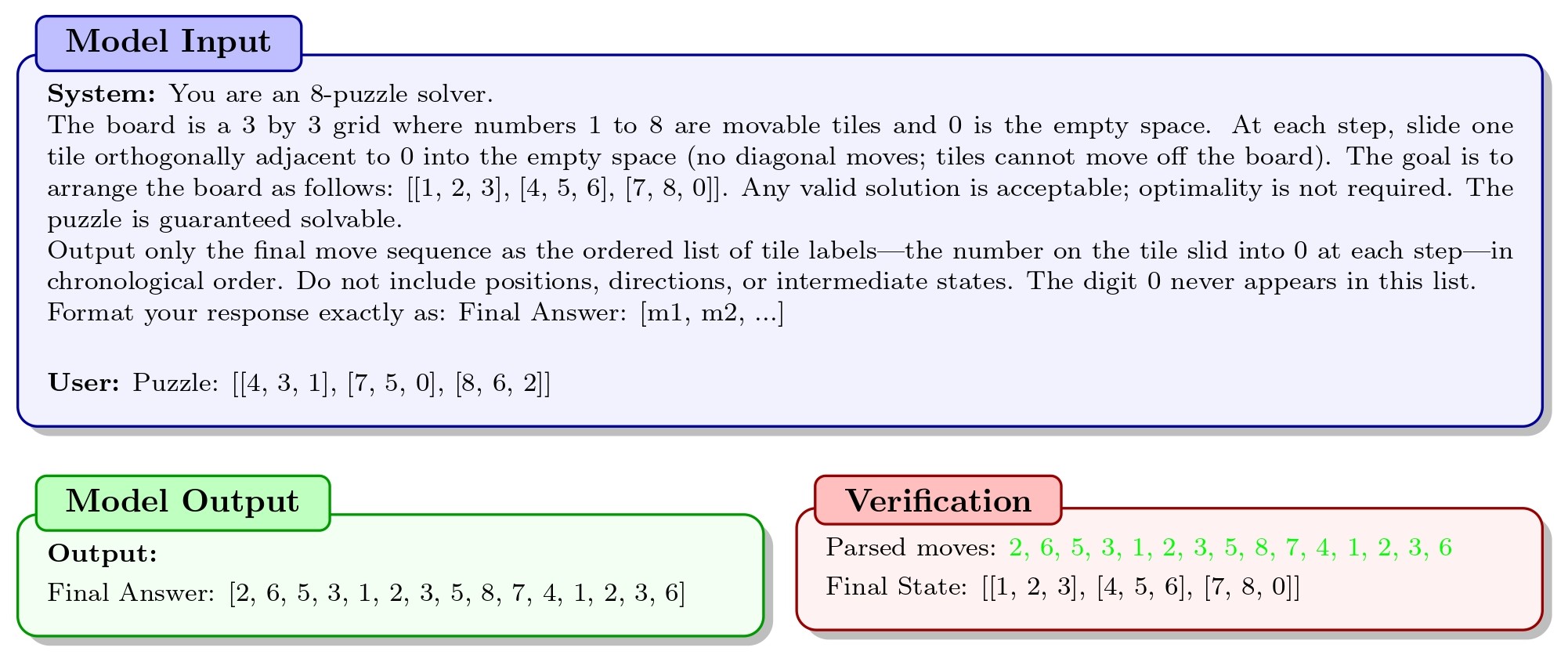}
  \caption{Example of our 8-puzzle evaluation pipeline. The model receives the system prompt, in this case, our Zero-Shot prompt, as well as the puzzle. The model then returns a sequence of moves. An external function applies each move to determine whether the model solved the puzzle. We highlight the legal moves in green. In this example, the model solves the puzzle on its first attempt.}
  \label{fig:pipeline}
\end{figure}

In recent years, Large Language Models (LLMs) have demonstrated state-of-the-art performance on an expanding range of tasks. Successive models exhibit increasingly sophisticated capabilities \cite{openai_gpt5_syscard_web_2025, gemini2025}, yet their evaluation still centers largely on mathematics and code generation benchmarks \cite{Hendrycks2020MMLU, balunovic2025matharena, Jimenez2023SWEBench}. This narrow focus leaves the dynamic, multi-step processes of reasoning and planning under-examined. When these abilities are studied directly, progress is often far less convincing than in other domains
\cite{Valmeekam2025ASystematicEvaluationofthePlanning, chollet2025arcprize, nezhurina2025aliceinwonder}. These limitations hinder the deployment of LLMs in complex, real-world applications that demand robust planning and state tracking, such as autonomous agents \cite{Liu2023AgentBench}.

Although LLMs can often solve such problems by writing or calling basic search code, our goal here is to isolate their intrinsic planning and state-tracking abilities without such tools. We therefore ask a concrete question: how well can LLMs plan and reason over an evolving state when they cannot rely on code execution or other external tools? To answer this, we use the standard 8-puzzle and ask models to solve random configurations of the game. The 8-puzzle requires models to maintain an accurate representation of the board, obey strict move validity constraints, and choose moves that eventually reach a goal state. We measure not only whether a puzzle is solved but also how and why runs terminate when they fail, to understand the limits of model performance on this type of task.

We investigate the performance of four models—GPT-5-Thinking, Gemini-2.5-Pro, GPT-5-mini, and Llama 3.1 8B-Instruct—on 8-puzzle instances with varying optimal solution lengths, using three common prompting strategies: Zero-Shot, Chain-of-Thought \cite{wei2022cot, yu2023bettercot}, and Algorithm-of-Thought \cite{sel2024aot}. Under each prompting strategy, each model receives a single attempt per puzzle. We then further assist the models on puzzles they initially fail to solve by allowing three additional attempts per level of feedback (repeat, specific, and suggestive). Within each feedback level, we save progress between calls. With these additional attempts, we observe more success under some conditions for particular models (GPT-5-Thinking with AoT and suggestive feedback). However, no single approach performs best across models, and the improvements are slow and costly. The final intervention we investigate is an external move validator that offloads the burden of determining move validity. The models no longer have to decide which moves are valid and which are not; we provide them with a list of all valid moves and ask them to return the best one. We then apply the move, update the board state and list, and reprompt the model. Despite this assistance, no model solves any puzzles.

Across models and conditions, two dominant deficits emerge: fallible representations of the board state, leading to invalid moves, and weak heuristic planning, leading to loops or moves that do not advance the puzzle toward the goal state. This work moves beyond aggregate performance metrics to offer a granular, qualitative analysis of why LLMs struggle with such tasks and how different interventions modulate these failures. Our contributions can be summarized as:

\begin{itemize}
\item A tool-free 8-puzzle evaluation across four LLMs under Zero-Shot, CoT and AoT prompting and tiered feedback, with a common protocol that isolates the effects of prompting versus feedback.

\item A fine-grained failure-mode analysis with per-condition breakdowns, revealing how and why models terminate instead of only reporting success rates.

\item An external move validator condition that eliminates the need to determine which moves are valid, isolating planning capabilities and quantifying each model's ability to work toward a goal when state tracking is no longer the bottleneck.
\end{itemize}

\section{Related Work}

\subsection{Improving LLM Reasoning and Planning}

Many studies have investigated how to improve LLM reasoning without fine-tuning models \cite{wei2022cot, wang2023selfconsistency, yao2023treeofthoughts, sel2024aot}. A common approach that consistently shows improvement is prompt engineering, where one uses the system prompt to specify how the model should approach a problem. Chain-of-Thought \cite{wei2022cot, wang2023understandingcot} is a familiar example, and it often asks the model to work step-by-step through a task. Algorithm-of-Thought \cite{sel2024aot} is another approach that instructs the model to think like a search algorithm. Other works seek to decompose the problem \cite{bi2025forestofthought, yao2023treeofthoughts}, allowing the model to explore several reasoning branches and then proceed with the most promising one based on a heuristic such as self-critique. Notably, even with guidance on how models should approach the problem, these methods still rely on the models' own capabilities to complete the task.

However, other work has offloaded some of the burden from the models when solving problems. For example, some works improve LLM performance by using tools such as code interpreters, which offload part of the computation to a separate program \cite{gao2023pal, chen2023pot}. Additional work improves model performance by enabling models to access the internet and consult resources such as Wikipedia \cite{schick2023toolformer, yao2023react}. With these extra tools, the models often outperform prompt-only methods. However, their performance can no longer be attributed solely to the models themselves, as it now depends on external information and computation.

\subsection{Evaluation Methods}

LLM progress is often measured based on benchmark performance, such as AIME \cite{MAA2025InvitationalCompetitions, balunovic2025matharena}, SWE-bench \cite{Jimenez2023SWEBench}, or others \cite{phan2025hle, wang2024mmlupro, kazemi2025bbeh}, highlighting their proficiency in mathematics, writing code, or other complex tasks. These benchmarks often rely on arithmetic or knowledge-based questions, where success is determined by recalling facts or using similar examples from the training data to solve the problems presented. However, many studies have examined the possibility of data contamination \cite{balunovic2025matharena, xu2024bdc, liang2025swebenchillusion}, raising concerns about the validity of the improvements we observe across models. Furthermore, these tasks offer less insight into the models' planning abilities because they rarely require a long-term strategy.

Work that focuses on LLM planning often reports low success rates and provides high-level analyses of why models fail—for example, prompt-sensitive pattern-matching and the need for planner-based repair \cite{Valmeekam2023OnThePlanningAbilitiesLLMs}. Other recent evaluations include cost/efficiency comparisons with some failure-mode breakdowns \cite{Valmeekam2025ASystematicEvaluationofthePlanning, shojaee2025illusion}, but they do not typically emphasize fine-grained analyses of how failures unfold. Furthermore, few studies examine variable levels of feedback, and we are aware of none that systematically combine modern prompting techniques with feedback on a planning task like the 8-puzzle to analyze how these factors modulate performance and failure types, leaving the effects of prompting and feedback on such tasks under-examined.

\section{Methods}

\subsection{8-puzzle}

The 8-puzzle is a simple and widely used search problem in AI \cite{russell2020aima, complete8puzzle}. The puzzle consists of a 3-by-3 grid with eight numbered, sliding tiles (1–8) and a single blank space (represented as 0). Tiles directly adjacent to the space can slide into it to change the board configuration. From a random starting configuration, the goal is to slide tiles until the board reads 1–8 from left to right, top to bottom, with the blank in the bottom-right corner. The 8-puzzle is particularly well-suited to our investigation because it is a well-studied problem in AI \cite{russell2020aima}, it provides a controllable and predictable environment in which the model must track the board state and obey strict rules, and it demands non-trivial long-term planning because solutions are unlikely to be found through local or random actions alone. Thus, it combines state tracking and planning in a single, compact domain, with easily verifiable solutions and unambiguous instructions.

The problem space consists of 181,440 unique solvable configurations, with the shortest optimal solution requiring zero moves and the longest optimal solution requiring 31 moves \cite{complete8puzzle}. To ensure fair representation of puzzles with varying difficulty, we divided them into difficulty bins based on their optimal solution lengths. Each bin represents roughly 20\% of the solvable puzzles and contains 10 sampled puzzles (see Fig.~\ref{fig:bins} for the bin definition). We generated the puzzles for each bin by creating a flattened list of integers from 0 to 8, shuffling them with Python’s \texttt{random.shuffle}, and reshaping them into 3-by-3 arrays. For each bin, we repeatedly sampled random boards and used only those that were solvable according to the standard parity test. Given the large number of possible puzzles and their random generation, we minimize the likelihood of data contamination. 

\subsection{Models}

We tested four models in our experiments: GPT-5-Thinking, Gemini-2.5-Pro, GPT-5-mini and Llama 3.1 8B-Instruct. To ensure uniformity and fairness between models, we used the same parameters and prompts for all four models: a token limit of 64,000, a time limit of 1,800 seconds per response, a temperature of 1.0, and identical system prompts. We set the temperature to 1.0 because the GPT-5 models do not allow variation from this value. Furthermore, a temperature of 1.0 is consistent with previous work \cite{shojaee2025illusion}, and even when variation is allowed, temperatures of 0.0 are not always deterministic with OpenAI models \cite{Valmeekam2025ASystematicEvaluationofthePlanning}. To control API costs, especially for the larger models, we evaluated all models on the same set of 50 puzzles, consistent with prior work \cite{shojaee2025illusion}.

\begin{figure}[t]
  \centering
  \includegraphics[width=1\linewidth]{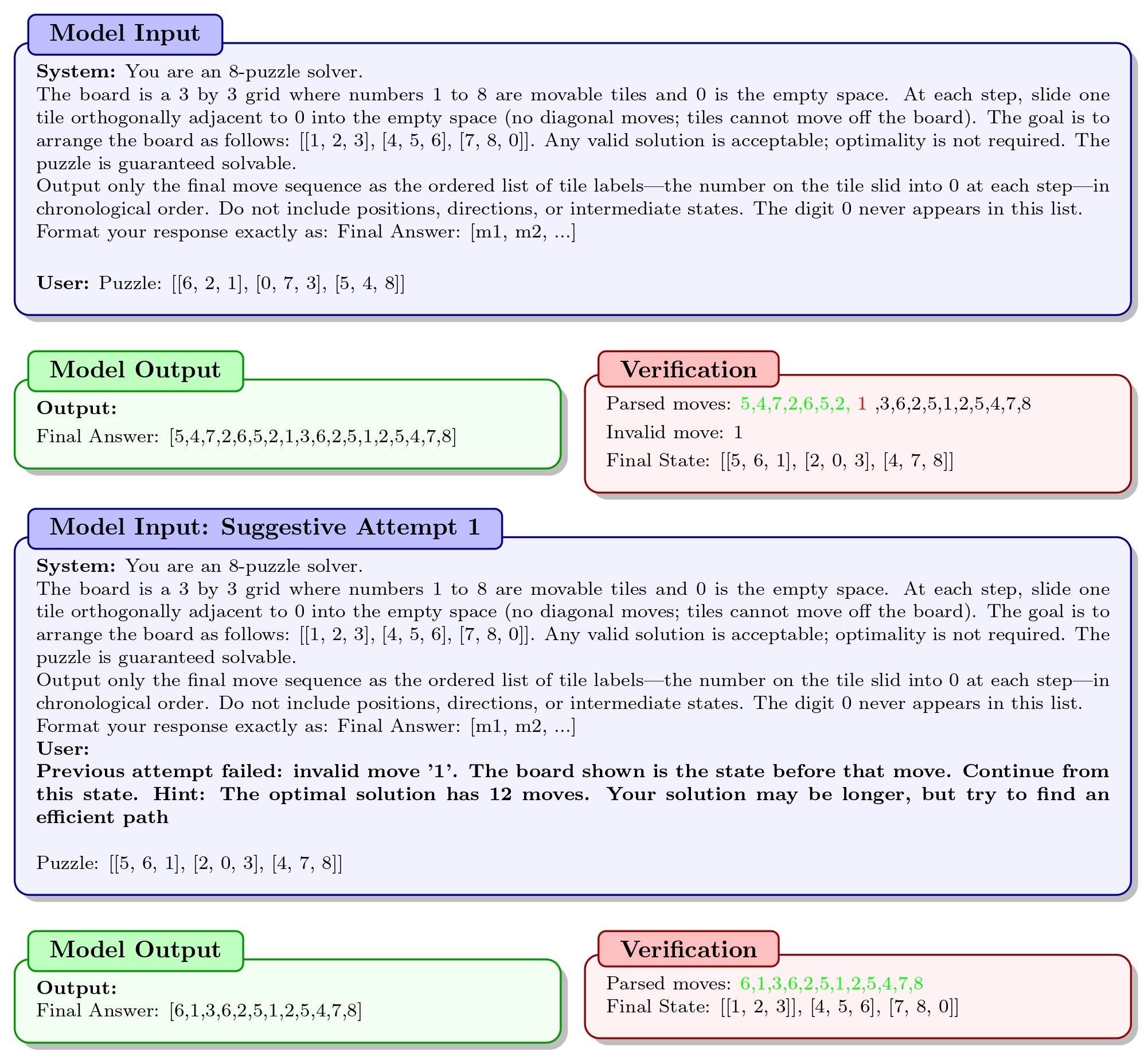}
  \caption{Example of our feedback pipeline. As in Fig.~\ref{fig:pipeline}, the model first receives the Zero-Shot system prompt and the puzzle, but its initial attempt fails due to an invalid move (highlighted in red). We go back to the preceding valid state and re-prompt the model with suggestive feedback (shown in bold), after which the model successfully solves the puzzle.}
  \label{fig:feedback_pipeline}
\end{figure}

\subsection{Prompting Strategies}

We study three prompting regimes under a common setup: \textbf{Zero-Shot} (no examples or reasoning cues), \textbf{CoT} (three worked examples encouraging stepwise reasoning), and \textbf{AoT} (three examples implementing a simple Manhattan-style search). The CoT and AoT prompts use the same three example boards, differing only in the reasoning styles they demonstrate. These three approaches serve as our \emph{base conditions}, on which we build all subsequent attempts with and without feedback.

\paragraph{Zero-Shot:} Our Zero-Shot prompt provides only the 8-puzzle rules, goal state, and output format, without examples or reasoning cues like "think step-by-step" \cite{kojima2022zeroshotcot}. This setup isolates the model's baseline reasoning and planning abilities and serves as a control to benchmark unguided performance against CoT and AoT interventions. We illustrate this setup in Fig.~\ref{fig:pipeline}.

\paragraph{Chain-of-Thought (CoT):} CoT is a prompting approach that often enhances LLM problem-solving by instructing the model to generate intermediate steps \cite{wei2022cot, yu2023bettercot, wang2023understandingcot}. We provided additional assistance using the few-shot paradigm, designing a 3-shot CoT prompt that gives the model a diverse set of worked examples. We chose three examples to limit prompt length and because prior work suggests that performance gains tend to diminish beyond a few examples \cite{wei2022cot, kojima2022zeroshotcot}. Our selected examples feature diverse board configurations and, critically, showcase complete solutions of varying lengths—3, 5, and 6 moves—and slightly different strategies to teach the model flexible and adaptive reasoning. The full prompt is available in Appendix \ref{sec:cot-prompt}.

\paragraph{Algorithm-of-Thought (AoT):} Algorithm-of-Thought works by guiding a language model to follow explicit algorithmic reasoning steps—often structured like pseudocode or formal rules—so that its outputs mimic systematic, step-by-step problem solving rather than free-form reasoning \cite{sel2024aot}. The original work reports improvements across many tasks and shows that AoT can outperform other prompting techniques, such as Chain-of-Thought (CoT) \cite{wei2022cot} and Tree-of-Thought (ToT) \cite{yao2023treeofthoughts, sel2024aot}. For our work, we specify a simple Manhattan-style search—similar to A*—and again provide the same three example boards to stay consistent with the CoT prompt and ensure a fair comparison between approaches. For further details on the AoT prompt, see Appendix \ref{sec:aot-prompt}.

\begin{figure}[t]
  \centering
  \includegraphics[width=1\linewidth]{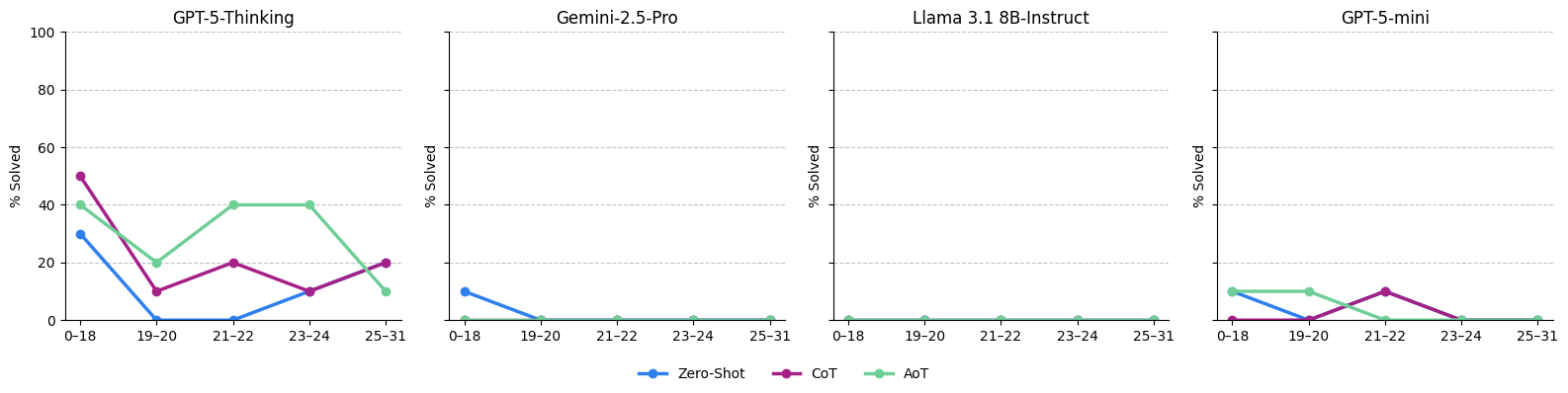}
  \caption{Success rates for each model under different prompting strategies across difficulty bins defined by optimal A* solution length. Each bin contains ten puzzles, and the percentage solved is computed within each bin.}
  \label{fig:bins}
\end{figure}

\subsection{Feedback} 

For each puzzle that a model failed to solve on its first attempt under any condition (Zero-Shot, CoT, or AoT), we ran three independent feedback trials: a repeat control, a trial with specific feedback, and one with suggestive feedback, as shown in Fig.~\ref{fig:feedback_pipeline}. All three feedback trials start from the same failure state and have up to three additional attempts to solve the puzzle. That is, within a given trial, the attempt \(k{+}1\) resumes from the \textbf{furthest valid state} reached in the attempt \(k\). Concretely, if model \(M\) failed puzzle \(i\) under AoT (even if it passed under Zero-Shot or CoT), we re-prompted \(M\) with the AoT system prompt for up to three attempts under each feedback condition: repeat (no feedback), specific feedback, and suggestive feedback.

By \emph{furthest valid state} we mean the last board configuration before failure: for \texttt{invalid\_move}, where the model moves a tile off the board or one that is not adjacent to the space, or \texttt{loop\_detected}, where the model repeats a board configuration, we resume from the state immediately before the offending move. If the model fails due to an \texttt{early\_stop}, where it makes only valid moves but stops before reaching the goal state, we resume from the state after the last move. We reset the visited-state set at the start of each attempt. Attempts stop early upon success; otherwise, they terminate due to a timeout, token limit, early stop, loop, invalid move, or graceful failure, in which the model either refuses to solve the problem or asks for clarification on how to do so. We cap retries at three per feedback condition (repeat, specific, and suggestive) to align with prior work on iterative LLM correction \cite{madaan2023selfrefine} and to limit API costs. Across feedback conditions, we keep the system prompt and hyperparameters fixed.

\paragraph{Repeat:} As a control condition, we re-prompt the model from the furthest valid state of its last failed attempt without providing any feedback. The prompt presents this state as a new, standalone puzzle. This trial serves as a baseline to measure performance gains arising purely from model stochasticity and from an additional attempt, thereby isolating the effects of the feedback provided under other conditions.

\paragraph{Specific:} Specific feedback also resumes from the furthest valid state, but provides precise details on the previous failed attempt. If the model makes an invalid move, we inform it of the error and identify the specific tile involved. If it exceeds the time limit, we specify the exact duration allowed. We designed this level to help the model avoid immediately repeating the same mistake and to highlight where it went wrong so it can correct its behaviour. For example, highlighting that the model made an invalid move should encourage it to focus more on tracking the state space.

\paragraph{Suggestive:} Suggestive feedback includes all elements of Specific feedback, plus a scalar hint: the \emph{optimal solution length} for the current board. We computed this value exactly using Iterative Deepening A* (IDA*) with the admissible Manhattan-distance heuristic. Because the Manhattan heuristic is admissible (and consistent) for the 8-puzzle, the returned length equals the true optimum. We reveal only the length \(x\)—never the action sequence—and append: ``Hint: The optimal solution has \(x\) moves. Your solution may be longer, but try to find an efficient path.'' The hint provides a progress signal while still leaving planning to the model, aiming to prevent it from undertaking an unnecessarily long search. We illustrate this level of feedback in Fig.~\ref{fig:feedback_pipeline}.

\subsection{External Move Validator}

In our final intervention, we offload part of the state-tracking burden from the model. The model receives the current puzzle state along with two lists: one containing all valid moves from the current position and another containing the previous move to help it avoid loops (e.g., immediately reversing the last action). This setup removes the need for the model to determine which moves are valid; it only has to examine the puzzle and select the best next move from the list. We instruct the model to return a single move, which we then apply to the puzzle. After applying the move, we update the puzzle state and move lists and re-query the model. Since the longest optimal solution for any 8-puzzle instance is 31 moves and roughly 94\% of puzzles require 26 or fewer moves, we allow the models to take up to 50 moves per puzzle, providing ample opportunity to solve it while avoiding excessively long, run-on trajectories.

\section{Results}

\subsection{LLMs Continue to Struggle}

\begin{figure}[t]
  \centering
  \includegraphics[width=0.90\linewidth]{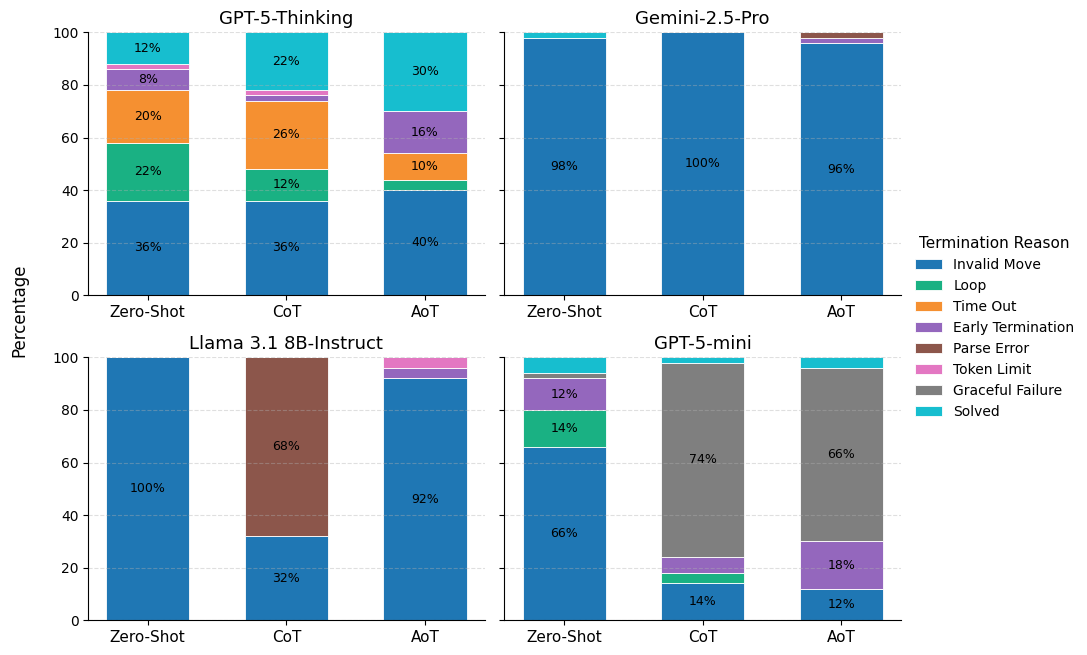}
  \caption{Termination breakdown of the four LLMs on the 8-puzzle across prompting strategies. Each stacked bar represents 100\% of trials for a given model and prompting condition (Zero-Shot, CoT, or AoT). The colored segments show the percentage of trials that ended in success (Solved) or in each failure mode.}
  \label{fig:main}
\end{figure}

Initial experiments established a performance baseline across all four models using Zero-Shot, CoT, and AoT prompting. The results, summarized in Fig.~\ref{fig:main}, reveal that successful puzzle completion was rare for most models. GPT-5-Thinking was the only model to achieve a modest success rate, with 30\% of puzzles solved under the AoT condition. This result represents the highest performance recorded in this phase, outperforming the next-best attempt (its own CoT performance) by eight percentage points.

In contrast to GPT-5-Thinking, the other models struggled significantly. Gemini-2.5-Pro and Llama 3.1 8B-Instruct were almost entirely unsuccessful, with Gemini solving only a single puzzle under the Zero-Shot condition and Llama solving none. The primary failure mode of these models is invalid moves across all prompting strategies, except Llama with CoT, which had a spike in parse errors because there was no list to extract from its answers. Interestingly, GPT-5-mini outperformed Gemini-2.5-Pro by solving more puzzles overall and at least one puzzle under each condition. GPT-5-mini also demonstrated a notable divergence from GPT-5-Thinking; its Zero-Shot performance was comparable to GPT-5-Thinking's at an 8\% success rate. However, unlike its more capable counterpart, introducing more complex CoT and AoT prompts did not improve performance, as its success rate dropped under both approaches. With GPT-5-mini, we also see, in both CoT and AoT, a sharp increase in graceful failure, where the model refuses to continue or requests clarification.

Reviewing success rates by difficulty, as shown in Fig.~\ref{fig:bins}, we observe that success does not decrease monotonically with optimal solution length. Although all models struggle with the longest puzzles, some achieve or match their best performance in the moderate-length bins. For example, GPT-5-Thinking with AoT matches its lower bin performance in two of the moderate bins. Similarly, GPT-5-mini, with its CoT prompt, solves no puzzles in the two shortest bins, then jumps up to 10\% in the middle one. This pattern points to planning and state-tracking issues, rather than puzzle difficulty alone, as the main determinants of success rates.

\subsection{Feedback Helps}

\begin{figure}[t]
  \centering
  \includegraphics[width=1\linewidth]{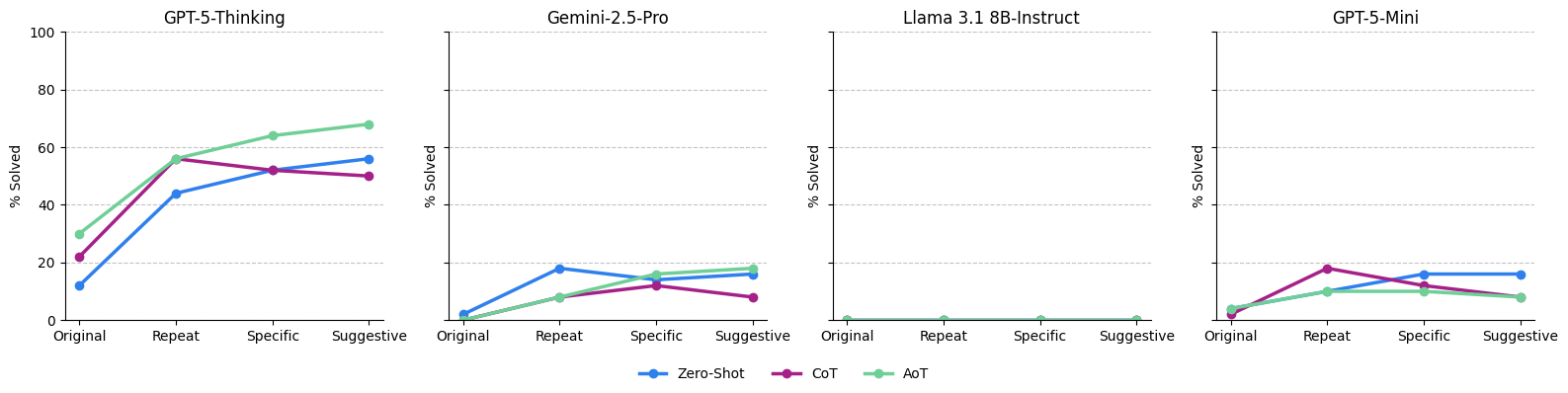}
  \caption{Success rates of each model under different feedback conditions. The “Original” point represents the initial success rate for each prompting strategy (Zero-Shot, CoT, and AoT). The “Repeat,” “Specific,” and “Suggestive” points show the final success rates after models were given up to three additional attempts on previously failed puzzles with the corresponding type of feedback. Each line tracks the performance of a prompting strategy across feedback conditions.}
  \label{fig:feedback}
\end{figure}

As shown in Fig.~\ref{fig:feedback}, providing feedback, additional attempts, and saving progress yielded performance gains across all models and prompting strategies, except for Llama 3.1 8B-Instruct. GPT-5-Thinking showed the most significant improvement, with AoT prompting and suggestive feedback yielding a 68\% success rate, the best performance for any model. As the level of feedback increased, GPT-5-Thinking's performance improved with AoT and Zero-Shot, but the opposite occurred with CoT: its performance steadily decreased. We observe the same trend for GPT-5-mini under CoT. These results suggest that more informative feedback may not always be desirable for these models.

The other models, excluding Llama 3.1 8B-Instruct, also showed steady gains. Both Gemini-2.5-Pro and GPT-5-mini achieved a peak success rate of 18\%. Notably, both models achieved this peak under the "Repeat" condition (Gemini-2.5-Pro with Zero-Shot and GPT-5-mini with CoT), suggesting that the primary benefit came from saving progress and re-prompting rather than from the feedback content itself. Gemini-2.5-Pro, however, matched this 18\% performance when using the combination of AoT prompting and suggestive feedback. GPT-5-mini with CoT, similar to GPT-5-Thinking with CoT, shows a steady decline in performance as the level of feedback increases, again highlighting that more informative feedback is not always beneficial. Finally, at the lowest end of the performance spectrum, Llama 3.1 8B-Instruct failed to solve any puzzles even with feedback, underscoring a significant capability gap on this task.

\subsection{External Move Validator Highlights Planning Deficits}

In the final experiment, we aimed to isolate planning abilities by offloading part of the state tracking responsibilities. The results, shown in Fig.~\ref{fig:outcome2}, reveal a critical failure in heuristic planning: none of the models solved any puzzles. The dominant failure mode for the models shifts to looping, in which they repeat a move and return to an already visited board configuration. GPT-5-Thinking looped in 100\% of trials, with Gemini-2.5-Pro and Llama 3.1 8B-Instruct looping 92\% and 86\% of the time, respectively. These results suggest that even when presented only with valid moves and the previous move to help track progress, these models struggle to formulate a goal-oriented path.

GPT-5-mini showed a different kind of limitation: it did not solve any puzzles, but its primary outcome was early termination (68\%), meaning it consistently made valid moves until it reached the 50-move limit. Therefore, GPT-5-mini exhibits a different kind of planning deficit: it can make valid moves and track the state space—contrasting with its earlier single-prompt results in Fig.~\ref{fig:main}—but lacks a coherent long-term strategy, effectively wandering the state space without direction.

\section{Discussion}

\subsection{What did we learn?}

Our experiments reveal that a diverse range of LLMs, differing in producer and size, continue to struggle with simple planning and search-based tasks such as the 8-puzzle, even SOTA models like GPT-5-Thinking, which has been described as possessing PhD-level intelligence \cite{openai_introducing_gpt5_2025}. Models consistently fail due to two primary deficits: brittle internal state representations that lead to invalid moves and weak heuristic planning that results in loops. These deficits indicate that models often fail to maintain accurate board state representations or to devise effective long-term strategies. This finding underscores a key limitation: while LLMs increasingly excel on academic benchmarks, they remain challenged by sequential tasks that demand sustained planning and state awareness.

The prevalence of loop failures is especially revealing because the system prompt explicitly instructs models to avoid repeating moves. Despite this explicit constraint, all models frequently violated the instruction, suggesting a fundamental difficulty in adhering to constraints. This deficit is most apparent in the external move validator experiment: even when given a list of valid moves, an explicit instruction not to repeat moves, and their previous move, the models continued to loop. One might argue that the model's internal heuristic changes between calls, but even then, it could simply select a different valid move when its preferred choice violates the constraints. Instead, it repeats moves, indicating that the planning process relies not only on weak heuristics but also fails to integrate the prompt's explicit rules.

\subsection{Effects of Prompting}

Our experiments suggest that the utility of advanced prompting strategies is model-dependent, as illustrated in Fig. \ref{fig:main}. While Algorithm-of-Thought (AoT) and Chain-of-Thought (CoT) improved performance for GPT-5-Thinking, they degraded performance for GPT-5-mini and Gemini-2.5-Pro. These findings demonstrate that, for this task, the efficacy of a prompting technique is contingent on the model, with no single approach proving universally superior.

The effects of prompting extended beyond success rates, causing significant shifts in termination modes (Fig. \ref{fig:main}). For instance, Llama 3.1 8B-Instruct under the CoT condition exhibited an increase in parse errors, indicating a failure to adhere to the specified output format. Concurrently, GPT-5-mini showed a dramatic increase in \texttt{graceful failure} with both CoT and AoT, in which the model refuses to provide a solution or requests clarification. These outcomes suggest that, for some models, introducing complex prompts creates confusion, hindering their ability to follow instructions rather than aiding their reasoning.

\subsection{The Cost of Feedback}

GPT-5-Thinking saw the most significant performance increase with the addition of feedback and saved progress between calls. Its success rate increased from 30\% to 68\%, but this comes at a substantial cost. As Table \ref{tab:table1} shows, the model needs, on average, two attempts, 24 minutes, over twice as many moves and 75,000 tokens to solve a puzzle. These averages are all higher when we include the puzzles that the models failed to solve. Furthermore, we obtain these results under carefully engineered prompts designed to be as helpful as possible for this task. Thus, while performance may increase, the compute and human resources required to provide these gains are substantial and problem-dependent. Gemini-2.5-Pro and GPT-5-mini both have similar issues with resource usage. While their average time, tokens and moves needed are lower overall, they require more attempts and have a substantially lower success rate of 18\%. 

\begin{table}[ht]
\centering
\footnotesize
\setlength{\tabcolsep}{3pt} 
\resizebox{1\textwidth}{!}{%
\begin{tabular}{lcccccccc}
\toprule
Model & Approach & Feedback & Solved N &
Time (min) &  Tokens & Attempts & Moves & Optimal \\
\midrule
GPT-5-Thinking        & AoT  & Suggestive & 34 & 23.92 & 75284 & 1.97 & 48.91  & 21.03 \\
Gemini-2.5-Pro        & Base & Repeat     &  9 &  6.74 & 58651 & 2.67 & 30.56 & 19.67 \\
Llama 3.1 8B-Instruct & Base & Repeat     &  0 &  0 &  0 &  0 & 0 & 0 \\
GPT-5-mini            & CoT  & Repeat     &  9 &  9.91 & 44272 & 2.67 & 28.44 & 19.11 \\
\bottomrule
\end{tabular}}
\caption{Solved count and average time (minutes), tokens used, attempts, moves, and optimal moves for each model under its best-performing condition. We broke ties for GPT-5-mini and Gemini-2.5-Pro by choosing the approach that required the fewest tokens.}
\label{tab:table1}
\end{table}

\begin{figure}[t]
  \centering
  \includegraphics[width=0.56\linewidth]{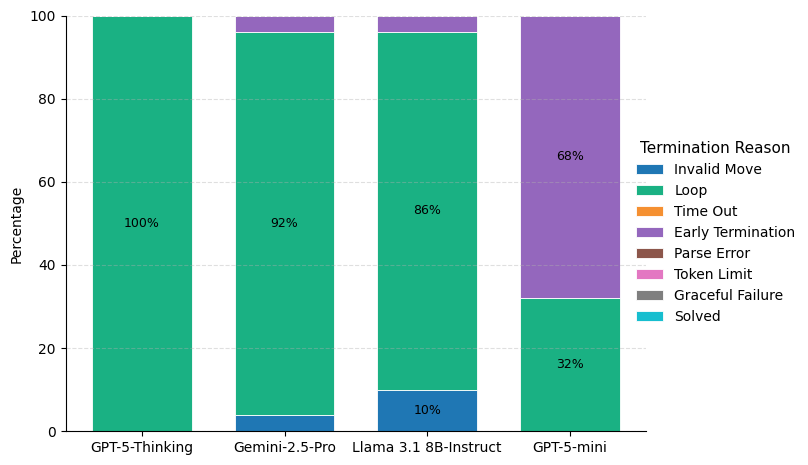}
  \caption{Termination breakdown of each LLM when using the external move validator. Each stacked bar represents 100\% of trials for a given model. The colored segments illustrate the percentage of trials that terminated due to each failure type. The color scheme is the same as the one used in Fig.  ~\ref{fig:main} }
  \label{fig:outcome2}
\end{figure}

\subsection{Planning with an External Move Validator}

\begin{figure}[t]
  \centering
  \includegraphics[width=0.55\linewidth]{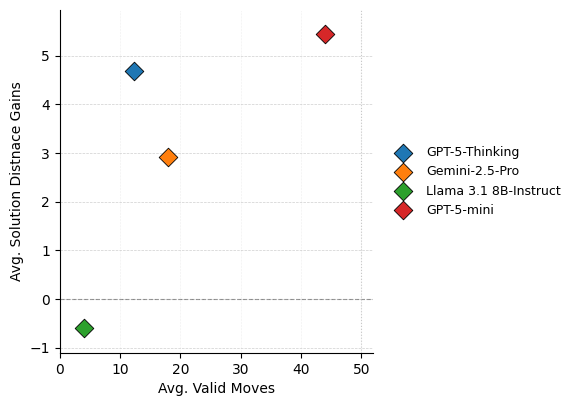}
  \caption{The average number of valid moves a model made with each puzzle vs the average distance gain from the solution based on the initial and final state using Manhattan distance.}
  \label{fig:external}
\end{figure}

In our final experiment with the external move validator, no model solved a single puzzle. As shown in Fig.~\ref{fig:outcome2}, this intervention hurt performance, lowering the success rates of all models and shifting the dominant failure mode to looping in most models and early termination in GPT-5-mini. Fig.~\ref {fig:external} further illustrates the planning deficit, showing that the models made minimal progress toward the goal state despite making numerous valid moves. For instance, GPT-5-mini made, on average, nearly forty-seven valid moves to achieve only about six moves of progress based on a Manhattan-distance heuristic. GPT-5-Thinking and Gemini were more efficient in that they made better progress in fewer moves, but this matters little, as they often began to repeat moves, completely stagnating progress. This finding highlights that, even when given only valid moves, models either rely on heuristics so weak that solving a typical puzzle could require well over 100 moves, or they begin repeating moves and cycling through already visited states.

\subsection{Implications}

Our findings, though derived from a single domain, carry significant implications for LLM deployment. We observe that all models struggle with this simple task, and while engineered prompts with feedback and saved-state progression show improvement, they are costly. These interventions also fail to make any model reliable, as none achieves a success rate over 70\%. The nature of these failures is also concerning, with the primary deficits not just weak heuristic planning that leads to loops, but also a persistent difficulty in maintaining an accurate representation of the board state, leading to "hallucinated" or invalid moves. These errors lead to a critical issue: models frequently output solutions as if they have correctly solved the puzzle, without detecting the mistakes made earlier in the sequence. This kind of false confidence undermines trust, because a model’s final output may rest on undetected errors earlier in the sequence.

These findings challenge the vision of deploying LLMs as autonomous agents in complex, real-world applications. If an LLM cannot reliably track its own state in a simple, deterministic puzzle, it may have similar issues in high-stakes, dynamic environments. In domains such as autonomous robotics, financial trading, or logistics management, an agent that proceeds with high confidence after an internal error can be unreliable and potentially dangerous. It risks costly or harmful outcomes by acting on a flawed premise.

\subsection{Absence of Coding Tools}

One might ask why we do not simply allow the models to use code interpreters for this task—after all, using tools is itself a form of reasoning. However, allowing the model to write and execute code changes the nature of the problem: the evaluation then focuses on whether the model can produce valid code, a capability already well studied by existing benchmarks. Moreover, code-based search largely masks planning and state tracking, since an external algorithm can solve the puzzle without the model needing to devise or maintain a long-term strategy. Therefore, in our experiments, we exclude code interpreters because they would mask the abilities we seek to examine.

\section{Conclusion}

In this work, we examined a range of LLMs using the 8-puzzle to probe their planning and stateful reasoning capabilities. Our findings show that both SOTA and small models still struggle significantly with this seemingly simple task, with failures indicating critical deficits in state tracking and heuristic planning. While feedback and prompt engineering offer some help, they are costly and do not provide the gains needed to make these models reliable. Furthermore, even when offloading some of the burden by providing the model with valid moves and its last move, no model solves any puzzles. The tendency for models to be unaware of their mistakes and to proceed with high confidence after undetected errors poses a substantial risk in real-world settings. These poor success rates underscore a fundamental gap between current LLM capabilities and the reliable, stateful reasoning required for genuine autonomy.

\section{Limitations}

Our work has several limitations. First, we examine only a single domain. While the 8-puzzle is informative for tracking a model's progress and may be representative of more challenging sequential planning tasks, it remains a single task, which limits our ability to draw broad conclusions about model performance in other settings. Second, we evaluate the models on only 50 puzzles, which is a relatively small dataset. Finally, although we generated puzzles at random to reduce the risk of training-data contamination, we cannot entirely rule it out.

\bibliography{references}     

\newpage

\appendix

\section{System Prompts}

\newtcolorbox{PromptBox}[1]{%
  breakable, enhanced,
  title=\textbf{#1},
  colback=gray!5, colframe=gray!60,
  boxrule=0.6pt, arc=1mm, left=6pt, right=6pt, top=6pt, bottom=6pt}

\subsection{Zero-Shot Prompt}
\label{sec:zero-prompt}

As outlined in our Methods section, we designed the Zero-Shot prompt to provide an overview of the problem and the rules the model must follow. We provide no additional assistance to the model. This format serves as our control condition, showing how strong each model is on its own, without prompting-based guidance.

\begin{PromptBox}{System Prompt — Zero-Shot}
You are an 8-puzzle solver.
\newline  
\space
\newline
The board is a 3 by 3 grid where numbers 1 to 8 are movable tiles and 0 is the empty space. At each step, slide one tile orthogonally adjacent to 0 into the empty space (no diagonal moves; tiles cannot move off the board). The goal is to arrange the board as follows: [[1, 2, 3], [4, 5, 6], [7, 8, 0]]. Any valid solution is acceptable; optimality is not required.  
\newline  
\space
\newline
The puzzle is guaranteed solvable. Output only the final move sequence as the ordered list of tile labels—the number on the tile slid into 0 at each step—in chronological order. 
Do not include positions, directions, or intermediate states. The digit 0 never appears in this list.
\newline  
\space
\newline
Format your response exactly as:
Final Answer: [m1, m2, ...]

\end{PromptBox}

\subsection{CoT Prompt}
\label{sec:cot-prompt}

We designed the few-shot CoT prompt to show the model several worked examples of how to solve different puzzles. The solutions vary in length but are kept relatively short to keep the prompt manageable. In each example, we explicitly track the state by repeating the current board after each move. This explicit state tracking is intended to encourage the model to follow a similar procedure and to minimize the number of invalid moves.

\begin{PromptBox}{System Prompt — CoT}

You are an 8-puzzle solver. 
\newline

The board is a 3 by 3 grid where numbers 1 to 8 are movable tiles and 0 is the empty space. At each step, slide one tile orthogonally adjacent to 0 into the empty space (no diagonal moves; tiles cannot move off the board). The goal is to arrange the board as follows: [[1, 2, 3], [4, 5, 6], [7, 8, 0]]. Any valid solution is acceptable; optimality is not required.  
\newline

The puzzle is guaranteed solvable. Think step-by-step to find the solution. After your reasoning, provide the final sequence of tile numbers that slide into the empty space as an array in the following format: Final Answer: [m1, m2, ...]
\newline

Here are three complete examples of how to solve a puzzle:
\newline

Example 1: 
\newline

User:\newline
Puzzle: [[1, 2, 3], [4, 8, 5], [7, 0, 6]]
\newline

Assistant: \newline
1. My goal is to get the board to the goal state. I will plan a sequence of moves. \newline
2. First, I need to get the '8' into its correct position. I will slide tile 8 down into the empty space. The board is now [[1, 2, 3], [4, 0, 5], [7, 8, 6]].\newline
3. Next, I need to get the '5' into its correct position. I will slide tile 5 left into the empty space. The board is now [[1, 2, 3], [4, 5, 0], [7, 8, 6]].\newline
4. Now, I can slide the '6' tile up into the empty space. The board becomes [[1, 2, 3], [4, 5, 6], [7, 8, 0]]. This is the goal state.\newline
5. My sequence of moves is [8, 5, 6].\newline

Output:\newline
Final Answer: [8, 5, 6]
\newline

Example 2:
\newline

User:\newline
Puzzle: [[1, 0, 3], [5, 2, 6], [4, 7, 8]]
\newline

Assistant:\newline
1. My goal is to get the board to the goal state. I will plan a sequence of moves.\newline
2. I see that I can get '2' into its correct position. I will slide 2 up into the empty space. The board is now [[1, 2, 3], [5, 0, 6], [4, 7, 8]].\newline
3. Now I can slide tile '5' right into its correct position. The board is now [[1, 2, 3], [0, 5, 6], [4, 7, 8]].\newline
4. Next, I need to get the '4' into its correct position. I will slide tile 4 up into the empty space. The board is now [[1, 2, 3], [4, 5, 6], [0, 7, 8]].\newline
5. Now I can slide the '7' tile left into its correct position. The board is now [[1, 2, 3], [4, 5, 6], [7, 0, 8]].\newline
6. I can now slide the '8' tile left into its correct position. The board is now [[1, 2, 3], [4, 5, 6], [7, 8, 0]]. This is the goal state.\newline
7. My sequence of moves is [2, 5, 4, 7, 8].\newline

Output: \newline
Final Answer: [2, 5, 4, 7, 8]
\newline

Example 3:
\newline

User: \newline
Puzzle: [[1, 3, 6], [4, 0, 2], [7, 5, 8]] \newline

Assistant: \newline
1. My goal is to get the board to the goal state. I will plan a sequence of moves. \newline
2. Reviewing the board configuration I see that if I move '2' left I can then move '6' into its correct location.\newline
3. Therefore, first I will slide tile '2' left into the empty space. The board is now [[1, 3, 6], [4, 2, 0], [7, 5, 8]].\newline
4. Now I can slide tile '6' down into its correct position. The board is now [[1, 3, 0], [4, 2, 6], [7, 5, 8]].\newline
5. Next, I can slide '3' right into its correct position. The board is now [[1, 0, 3], [4, 2, 6], [7, 5, 8]].\newline
6. Now I can slide '2' up into its correct position. The board is now [[1, 2, 3], [4, 0, 6], [7, 5, 8]].\newline
7. Now I will slide '5' up into its correct position. The board is now [[1, 2, 3], [4, 5, 6], [7, 0, 8]].\newline
8. I can now slide '8' left into its correct position. The board is now [[1, 2, 3], [4, 5, 6], [7, 8, 0]]. This is the goal state. \newline
9. My sequence of moves is [2, 6, 3, 2, 5, 8].\newline

Output:\newline
Final Answer: [2, 6, 3, 2, 5, 8]
\newline

Begin your step-by-step thinking now.

\end{PromptBox}

\subsection{AoT Prompt}
\label{sec:aot-prompt}

The few-shot AoT prompt is our longest system prompt. It contains the same three examples used in the CoT prompt, but the search-based procedure we demonstrate makes the prompt much longer in practice. It provides an explicit heuristic for the model to follow, and, as in the CoT prompt, we also illustrate state tracking step by step in each example.

\begin{PromptBox}{System Prompt — AoT}

You are an 8-puzzle solver.
\newline\newline
The board is a 3 by 3 grid where numbers 1 to 8 are movable tiles and 0 is the empty space. At each step, slide one tile orthogonally adjacent to 0 into the empty space (no diagonal moves; tiles cannot move off the board). The goal is to arrange the board as follows: [[1, 2, 3], [4, 5, 6], [7, 8, 0]]. Any valid solution is acceptable; optimality is not required.    
\newline\newline
The puzzle is guaranteed solvable. Think step-by-step to find the solution. After your reasoning, provide the final sequence of tile numbers that slide into the empty space as an array in the following format: Final Answer: [m1, m2, ...]
\newline\newline
You will use a Manhattan-guided search (A*-style scoring) to guide your thinking, prioritizing paths with a lower `f(s) = g(s) + h(s)`, where `g(s)` is the number of moves made and `h(s)` is the sum of Manhattan distances of tiles from their goal positions. Do not immediately backtrack, and keep a small set of seen boards to skip repeats. Show your search process by exploring promising moves and tracking your path. 
\newline\newline
Example 1:
\newline\newline
User:
\newline
Puzzle: [[1, 2, 3], [4, 8, 5], [7, 0, 6]]
\newline\newline
Assistant:
\newline
Initial State: [[1, 2, 3], [4, 8, 5], [7, 0, 6]]. h=3. 
\newline
Search Process: Empty space is at (2, 1). Possible moves are 8, 7, 6.
\newline

Move 8: h=2 -> Better
\newline
Move 7: h=4 -> Worse
\newline
Move 6: h=4 -> Worse
\newline

Move tile 8. Current path: [8].
State: [[1, 2, 3], [4, 0, 5], [7, 8, 6]], h=2. 
\newline
Search Process: Empty space is at (1, 1). Possible moves are 2, 4, 5, 8
\newline

Move 2: h=3 -> Worse \newline
Move 4: h=3 -> Worse \newline
Move 5: h=1 -> Better \newline
Move 8: h=3 -> Worse \newline

Move tile 5. Current path: [8, 5].
State: [[1, 2, 3], [4, 5, 0], [7, 8, 6]], h=1. \newline
Search Process: Empty space is at (1, 2). Possible moves are 3, 5, 6
\newline

Move 3: h=2 -> Worse\newline
Move 5: h=2 -> Worse\newline
Move 6: h=0 -> Better\newline

Move tile 6. Current path: [8, 5, 6].\newline
State: [[1, 2, 3], [4, 5, 6], [7, 8, 0]]\newline
Puzzle Solved!
\newline\newline
Output:
Final Answer: [8, 5, 6]
\newline\newline

Example 2:
\newline\newline
User:\newline
Puzzle: [[1, 0, 3], [5, 2, 6], [4, 7, 8]]
\newline\newline
Assistant:\newline
Initial State: [[1, 0, 3], [5, 2, 6], [4, 7, 8]]. h=5.\newline
Search Process: Empty space is at (0, 1). Possible moves are 1, 2, 3
\newline

Move 1: h=6 -> Worse \newline
Move 2: h=4 -> Better \newline
Move 3: h=6 -> Worse \newline

Move tile 2. Current path: [2].
State: [[1, 2, 3], [5, 0, 6], [4, 7, 8]], h=4. \newline
Search Process: Empty space is at (1, 1). Possible moves are 2, 5, 7, 6
\newline

Move 2: h=5 -> Worse \newline
Move 5: h=3 -> Better \newline
Move 7: h=5 -> Worse \newline
Move 6: h=5 -> Worse \newline

Move tile 5. Current path: [2, 5].
State: [[1, 2, 3], [0, 5, 6], [4, 7, 8]], h=3. \newline
Search Process: Empty space is at (1, 0). Possible moves are 1, 5, 4
\newline

Move 1: h=4 -> Worse\newline
Move 4: h=2 -> Better\newline
Move 5: h=4 -> Worse\newline

Move tile 4. Current path: [2, 5, 4].\newline
State: [[1, 2, 3], [4, 5, 6], [0, 7, 8]]\newline
Search Process: Empty space is at (2, 0). Possible moves are 4, 7\newline

Move 4: h=3 -> Worse\newline
Move 7: h=2 -> Better\newline

Move tile 7. Current path: [2, 5, 4, 7].\newline
State: [[1, 2, 3], [4, 5, 6], [7, 0, 8]]\newline
Search Process: Empty space is at (2, 1). Possible moves are 7, 6, 8
\newline

Move 7: h=2 -> Worse\newline
Move 6: h=2 -> Worse\newline
Move 8: h=0 -> Better\newline

Move tile 8. Current path: [2, 5, 4, 7, 8].\newline
State: [[1, 2, 3], [4, 5, 6], [7, 8, 0]]\newline
Puzzle Solved!
\newline\newline
Output:\newline
Final Answer: [2, 5, 4, 7, 8]
\newline\newline
Example 3:
\newline\newline
User:\newline
Puzzle: [[1, 3, 6], [4, 0, 2], [7, 5, 8]]
\newline\newline
Assistant:\newline
Initial State: [[1, 3, 6], [4, 0, 2], [7, 5, 8]]. h=6.\newline
Search Process: Empty space is at (1, 1). Possible moves are 3, 4, 5, 2.
\newline

Move 3: h=7 -> Worse \newline
Move 4: h=7 -> Worse \newline
Move 5: h=5 -> Better \newline
Move 2: h=5 -> Better \newline

Two moves give equal gains. I'll try move 2 first. \newline
If in three moves I am not making progress I will come back to this position and try move 5. \newline

Move tile 2. Current path: [2].
State: [[1, 3, 6], [4, 2, 0], [7, 5, 8]]. h=5. \newline
Search Process: Empty space is at (1, 2). Possible moves are 6, 2, 8. \newline

Move 6: h=5 -> Better \newline
Move 2: h=6 -> Worse \newline
Move 8: h=6 -> Worse \newline

Move tile 6. Current path: [2, 6].
State: [[1, 3, 0], [4, 2, 6], [7, 5, 8]]. h=4. \newline
Search Process: Empty space is at (0, 2). Possible moves are 3, 6.
\newline

Move 3: h=3 -> Better \newline
Move 6: h=5 -> Worse [skip: visited/immediate backtrack] \newline

Move tile 3. Current path: [2, 6, 3]. \newline
State: [[1, 0, 3], [4, 2, 6], [7, 5, 8]]. h=3. \newline
Search Process: Empty space is at (0, 1). Possible moves are 1, 2, 3.
\newline

Move 1: h=4 -> Worse \newline
Move 2: h=2 -> Better \newline
Move 3: h=4 -> Worse \newline

Move tile 2. Current path: [2, 6, 3, 2]. \newline
State: [[1, 2, 3], [4, 0, 6], [7, 5, 8]]. h=2. \newline
Search Process: Empty space is at (1, 1). Possible moves are 2, 4, 6, 5. \newline

Move 2: h=3 -> Worse \newline
Move 4: h=3 -> Worse \newline
Move 5: h=1 -> Better \newline
Move 6: h=3 -> Worse \newline

Move tile 5. Current path: [2, 6, 3, 2, 5]. \newline
State: [[1, 2, 3], [4, 5, 6], [7, 0, 8]]. h=1. \newline
Search Process: Empty space is at (2, 1). Possible moves are 5, 7, 8. \newline

Move 5: h=2 -> Worse \newline
Move 7: h=2 -> Worse \newline
Move 8: h=0 -> Better \newline

Move tile 8. Current path: [2, 6, 3, 2, 5, 8]. \newline
State: [[1, 2, 3], [4, 5, 6], [7, 8, 0]] \newline 
Puzzle Solved!
\newline \newline
Output: \newline
Final Answer: [2, 6, 3, 2, 5, 8]
\newline

Begin your step-by-step thinking now.
\end{PromptBox}

\subsection{External Move Validator}
\label{sec:ext-prompt}

The external move validator prompt is similar to the Zero-Shot prompt; however, we now provide a list of valid moves from the current state and the previous move to prevent the model from immediately repeating a move. Aside from the variation in state, valid moves, and the previous move, we pass the same prompt to the model on each call.

\begin{PromptBox}{System Prompt — External Move Validator}

You are an 8-puzzle solver. 
\newline \newline
The board is a 3 by 3 grid where numbers 1 to 8 are movable tiles and 0 is the empty space.
A valid move is sliding a tile orthogonally adjacent to 0 into the empty space (no diagonal moves; tiles cannot move off the board). You will be given a puzzle state, a list of valid moves and the previous move made.
Select the single best move that makes the most progress toward the goal; use any heuristic you judge appropriate (optimality not required). If multiple moves are equally good, you may choose any of them but do not undo the previous move. The goal is to arrange the board as follows: [[1, 2, 3], [4, 5, 6], [7, 8, 0]].
\newline

The puzzle is guaranteed solvable.
Your response must be only the single integer representing your chosen move. Do not include any other words, explanations, or formatting.
\newline

Example:
\newline
\newline
Input:
\newline
Puzzle: [[1, 2, 3], [4, 5, 6], [0, 7, 8]]
\newline
Previous move: 4
\newline
Valid moves: [4, 7]
\newline
\newline
Output:
\newline
7

\end{PromptBox}

\section{Feedback}

\subsection{Specific Feedback}

With specific feedback we tell the model exactly why it failed. We provide specific limits on time and tokens. We also tell it the exact move it made that caused it to fail in the case of invalid moves, early stops and loops. 

\begin{PromptBox}{Exact Feedback Given}

PARSE\_ERROR = "Previous attempt failed due to invalid formatting of the final answer. Try again."
\newline

NO\_REPLY = "Previous attempt produced no response. Try again."
\newline

TIMEOUT = "Previous attempt exceeded the 1800-second limit. Try again."
\newline

TOKEN\_LIMIT = "Previous attempt exceeded the 64,000-token limit. Try again."
\newline

GRACEFUL\_FAILURE = "Previous attempt ended without a move sequence. This puzzle is solvable within the established constraints (1800-second limit and 64,000-token limit). Try again."
\newline

INVALID\_MOVE = "Previous attempt failed: invalid move \textit{x} (the tile number). The board shown is the state before that move. Continue from this state. "
\newline

EARLY\_STOP = "Previous attempt ended early. Your last valid move was \textit{x} (the tile number). The board shown is the state after that move. Continue from this state. "
\newline

LOOP\_DETECTED = "Previous attempt created a loop after \textit{x} (the tile number). The board shown is the state before that move. Continue from this state. "

\end{PromptBox}

\subsection{Suggestive Feedback}

Suggestive feedback uses the same prompts as specific feedback; the only modification is that we append a hint to the end of the feedback. This hint specifies the length of the optimal solution from current state.

\begin{PromptBox}{Example of Suggestive Feedback for Timeout Error}

Previous attempt exceeded the 1800-second limit. Try again. \newline
Hint: The optimal solution has  \textit{x} moves. Your solution may be longer, but try to find an efficient path.

\end{PromptBox}

\section{Full Worked Examples}

\subsection{GPT-5-Mini Zero-Shot with Specific Feedback}

\begin{PromptBox}{First Call to Model}
System:
\newline
Zero-Shot system prompt (see Appendix~\ref{sec:zero-prompt}).\newline  

User:\newline
Puzzle: [[1, 8, 0], [5, 6, 2], [4, 7, 3]]
\newline

Output:\newline
Final Answer: [2, 3, 7, 6, 8, 2, 3, 8, 5, 4, 6, 7, 8, 6, 7, 8]

\end{PromptBox}

\begin{PromptBox}{Compute Moves}

After thirteen legal moves, the model then proposes the invalid move 6 from the following state: \newline
[[1, 2, 3], [4, 5, 0], [6, 7, 8]]

\end{PromptBox}

\begin{PromptBox}{Retry Attempt 1}
System:
\newline
Zero-Shot system prompt (see Appendix~\ref{sec:zero-prompt}).\newline 

User:\newline
Previous attempt failed: invalid move 6. The board shown is the state before that move. Continue from this state.\newline
Puzzle: [[1, 2, 3], [4, 5, 0], [6, 7, 8]]
\newline

Output:\newline
Final Answer: [8, 7, 6, 4, 5, 6, 8, 6, 5, 4, 8]

\end{PromptBox}

\begin{PromptBox}{Compute Moves}

After six legal moves, the model then proposes the invalid move 8 from the following state: \newline
[[1, 2, 3], [5, 6, 8], [4, 0, 7]]

\end{PromptBox}

\begin{PromptBox}{Retry Attempt 2}
System:
\newline
Zero-Shot system prompt (see Appendix~\ref{sec:zero-prompt}).\newline 

User:\newline
Previous attempt failed: invalid move 8. The board shown is the state before that move. Continue from this state.\newline
Puzzle: [[1, 2, 3], [5, 6, 8], [4, 0, 7]]
\newline

Output:\newline
Final Answer: [7, 8, 6, 5, 4, 7, 8]

\end{PromptBox}

\begin{PromptBox}{Compute Moves}

After applying the moves the model proposes, we reach the goal state: \newline
[[1, 2, 3], [4, 5, 6], [7, 8, 0]].

\end{PromptBox}

\subsection{Gemini-2.5-Pro CoT with Repeat}

\begin{PromptBox}{First Call to Model}
System:
\newline
CoT system prompt (see Appendix~\ref{sec:cot-prompt}).\newline

User:\newline
Puzzle: [[6, 2, 1], [0, 7, 3], [5, 4, 8]]
\newline

Output:\newline
1. My goal is to solve the puzzle by getting the tiles into their correct positions. I will work on solving the puzzle row by row, starting with the top row.
\newline

2. The initial board is [[6, 2, 1], [0, 7, 3], [5, 4, 8]]. First, I want to get the numbers 1, 2, and 3 into the top row.
\newline

3. I'll start by moving the empty space to the top right to get the '3' into place. I slide tile '6' right. The board is now [[0, 2, 1], [6, 7, 3], [5, 4, 8]].\newline

4. Slide '2' right. The board is now [[2, 0, 1], [6, 7, 3], [5, 4, 8]].
\newline

5. Slide '1' left. The board is now [[2, 1, 0], [6, 7, 3], [5, 4, 8]].
\newline

6. Slide '3' up. The board is now [[2, 1, 3], [6, 7, 0], [5, 4, 8]]. The '3' is now in its final position.
\newline

7. Now I will work on getting '1' and '2' into position without disturbing '3'. I slide '7' right, then '4' up, then '5' right, then '6' down. This moves the empty space to the top left corner. The intermediate boards are:
\newline
   - Move 7: [[2, 1, 3], [6, 0, 7], [5, 4, 8]]
   \newline
   - Move 4: [[2, 1, 3], [6, 4, 7], [5, 0, 8]]
   \newline
   - Move 5: [[2, 1, 3], [6, 4, 7], [0, 5, 8]]
   \newline
   - Move 6: [[0, 1, 3], [2, 4, 7], [6, 5, 8]]
\newline

8. Now I can place '1' into its final spot. I slide '1' right. The board is now [[1, 0, 3], [2, 4, 7], [6, 5, 8]].
\newline

9. Now I can place '2' into its final spot. I slide '2' up. The board is now [[1, 2, 3], [0, 4, 7], [6, 5, 8]]. The first row is now solved.
\newline

10. My next goal is to solve the second row to be [4, 5, 6]. I start by sliding tile '4' left into the empty space. The board becomes [[1, 2, 3], [4, 0, 7], [6, 5, 8]]. The '4' is now in place.
\newline

11. Next, I slide '5' up. The board is now [[1, 2, 3], [4, 5, 7], [6, 0, 8]]. The '5' is now in place.
12. To get '6' into its final position, I will rotate the remaining pieces. I slide '6' right. The board is [[1, 2, 3], [4, 5, 7], [0, 6, 8]].
\newline

13. I slide '7' down. The board is [[1, 2, 3], [4, 5, 0], [7, 6, 8]].
\newline

14. I slide '6' up. The board is [[1, 2, 3], [4, 5, 6], [7, 0, 8]]. The second row is now solved.
\newline

15. Finally, I need to solve the last row. I need to swap the '0' and the '8'. I slide '8' left into the empty space. The board becomes [[1, 2, 3], [4, 5, 6], [7, 8, 0]]. This is the goal state.
\newline

16. My sequence of moves is [6, 2, 1, 3, 7, 4, 5, 6, 1, 2, 4, 5, 6, 7, 6, 8].
\newline

Final Answer: [6, 2, 1, 3, 7, 4, 5, 6, 1, 2, 4, 5, 6, 7, 6, 8]

\end{PromptBox}

\begin{PromptBox}{Compute Moves}

After eight legal moves, the model then proposes the invalid move 1 from the following state: \newline
[[2, 1, 3], [0, 4, 7], [6, 5, 8]]

\end{PromptBox}

\begin{PromptBox}{Retry Attempt 1}
System:
\newline
CoT system prompt (see Appendix~\ref{sec:cot-prompt}).\newline

User:\newline
Puzzle: [[6, 2, 1], [0, 7, 3], [5, 4, 8]]
\newline

Output:\newline
1. My goal is to get the board to the goal state `[[1, 2, 3], [4, 5, 6], [7, 8, 0]]`. I will solve the puzzle by first getting the top row in order, then the first column, and then the remaining tiles.\newline

2. The initial board is `[[2, 1, 3], [0, 4, 7], [6, 5, 8]]`. Tiles '1' and '2' are swapped. I will slide tile '2' down into the empty space. The board becomes `[[0, 1, 3], [2, 4, 7], [6, 5, 8]]`.\newline

3. Now the empty space is in the top-left corner. I can slide tile '1' left into its correct position. The board becomes `[[1, 0, 3], [2, 4, 7], [6, 5, 8]]`.\newline

4. Next, I will slide tile '2' up into its correct position. The board is now `[[1, 2, 3], [0, 4, 7], [6, 5, 8]]`. The first row is now correct.\newline

5. Now I will work on the first column. I need tile '4' in position (1, 0). The empty space is there, so I will slide tile '4' left into its correct place. The board becomes `[[1, 2, 3], [4, 0, 7], [6, 5, 8]]`.\newline

6. Next, I need to get tile '7' into position (2, 0). To do this without disturbing the solved tiles, I will maneuver the empty space to position (2, 0) and then slide '7' down.
   - First, slide tile '7' left: `[[1, 2, 3], [4, 7, 0], [6, 5, 8]]`.
   - Then, slide tile '8' left: `[[1, 2, 3], [4, 7, 8], [6, 5, 0]]`.
   - Slide tile '5' right: `[[1, 2, 3], [4, 7, 8], [6, 0, 5]]`.
   - Slide tile '6' right: `[[1, 2, 3], [4, 7, 8], [0, 6, 5]]`.
   - Now the empty space is at (2,0). Slide tile '7' down: `[[1, 2, 3], [4, 0, 8], [7, 6, 5]]`. The first column is now correct.\newline

7. Now I need to solve the remaining 2x2 block. The target is to have the second row as `[4, 5, 6]`. I will move the tiles `5, 6, 8` around the empty space to arrange them.
   - Slide '6' up: `[[1, 2, 3], [4, 6, 8], [7, 0, 5]]`.
   - Slide '5' left: `[[1, 2, 3], [4, 6, 8], [7, 5, 0]]`.
   - Slide '8' down: `[[1, 2, 3], [4, 6, 0], [7, 5, 8]]`.
   - Slide '6' right: `[[1, 2, 3], [4, 0, 6], [7, 5, 8]]`.
   - Slide '5' up: `[[1, 2, 3], [4, 5, 6], [7, 0, 8]]`. The second row is now correct.\newline

8. Finally, I only need to arrange the last row. The board is `[[1, 2, 3], [4, 5, 6], [7, 0, 8]]`. I will slide tile '8' left into the empty space. The board becomes `[[1, 2, 3], [4, 5, 6], [7, 8, 0]]`, which is the goal state.\newline

9. My sequence of moves is [2, 1, 2, 4, 7, 8, 5, 6, 7, 6, 5, 8, 6, 5, 8].\newline

Final Answer: [2, 1, 2, 4, 7, 8, 5, 6, 7, 6, 5, 8, 6, 5, 8]

\end{PromptBox}

\begin{PromptBox}{Compute Moves}

After two legal moves, the model then proposes the invalid move 2 from the following state: \newline
[[1, 0, 3], [2, 4, 7], [6, 5, 8]]

\end{PromptBox}

\begin{PromptBox}{Retry Attempt 2}
System:\newline
CoT system prompt (see Appendix~\ref{sec:cot-prompt}).\newline

User:\newline
Puzzle: [[6, 2, 1], [0, 7, 3], [5, 4, 8]]
\newline

Output:\newline
1. My goal is to get the board to the goal state `[[1, 2, 3], [4, 5, 6], [7, 8, 0]]`. I will plan a sequence of moves.\newline

2. The initial board is `[[1, 0, 3], [2, 4, 7], [6, 5, 8]]`. The 1 and 3 are in the correct place. I see I can move tile '2' up into the empty spot to complete the first row.\newline

3. Slide tile '2' up. The board is now `[[1, 2, 3], [0, 4, 7], [6, 5, 8]]`. The sequence of moves is `[2]`.\newline

4. Now the top row is correct. I will work on the second row, which should be `[4, 5, 6]`. The empty space is in the correct spot for tile '4'. \newline

5. Slide tile '4' left. The board is now `[[1, 2, 3], [4, 0, 7], [6, 5, 8]]`. The moves are `[2, 4]`.\newline

6. I need to get '5' and '6' into the second row. Let's move tile '6' up into the empty space.\newline

7. Slide tile '6' up. The board is now `[[1, 2, 3], [4, 6, 7], [0, 5, 8]]`. The moves are `[2, 4, 6]`.\newline

8. Now I need to maneuver the remaining tiles to get the second row correct. I will move the empty space next to tile '7'. \newline

9. Slide tile '5' right. The board is `[[1, 2, 3], [4, 6, 7], [5, 0, 8]]`. Moves: `[2, 4, 6, 5]`.\newline

10. Slide tile '8' left. The board is `[[1, 2, 3], [4, 6, 7], [5, 8, 0]]`. Moves: `[2, 4, 6, 5, 8]`.\newline

11. Now the empty space is next to '7'. Slide tile '7' down. The board is `[[1, 2, 3], [4, 6, 0], [5, 8, 7]]`. Moves: `[2, 4, 6, 5, 8, 7]`.\newline

12. The empty space is now at (1,2), where '6' should go. But '6' is at (1,1). I will now get '5' into its correct spot at (1,1).\newline

13. Slide tile '6' left. The board becomes `[[1, 2, 3], [4, 0, 6], [5, 8, 7]]`. Moves: `[2, 4, 6, 5, 8, 7, 6]`.\newline

14. Now the empty space is at (1,1), where '5' should be. Tile '5' is directly below it.\newline

15. Slide tile '5' up. The board is now `[[1, 2, 3], [4, 5, 6], [0, 8, 7]]`. The first two rows are correct. Moves: `[2, 4, 6, 5, 8, 7, 6, 5]`.
16. The last row is `[0, 8, 7]`. I need to swap '8' and '7'. I will move '8' into the empty spot.\newline

17. Slide tile '8' right. The board is `[[1, 2, 3], [4, 5, 6], [8, 0, 7]]`. Moves: `[..., 5, 8]`.\newline

18. Slide tile '7' left. The board is `[[1, 2, 3], [4, 5, 6], [8, 7, 0]]`. Moves: `[..., 8, 7]`.\newline

19. Now the empty space is in its final position, but '7' and '8' are swapped. I will perform a three-tile rotation using tile '6'.
20. Slide tile '6' down. The board is `[[1, 2, 3], [4, 5, 0], [8, 7, 6]]`. Moves: `[..., 7, 6]`.\newline

21. Slide tile '7' right. The board is `[[1, 2, 3], [4, 5, 7], [8, 0, 6]]`. Moves: `[..., 6, 7]`.\newline

22. Slide tile '8' left. The board is `[[1, 2, 3], [4, 5, 7], [0, 8, 6]]`. Moves: `[..., 7, 8]`.\newline 

23. Slide tile '4' down is not allowed. I will slide `8` right. The board is `[[1, 2, 3], [4, 5, 7], [8, 0, 6]]`. This is a previous state. I will use a different procedure.

Let's try a simpler sequence from `[[1, 2, 3], [4, 5, 6], [0, 8, 7]]`.
1. Slide tile '8' right into the empty space. Board is `[[1, 2, 3], [4, 5, 6], [8, 0, 7]]`.\newline
2. Slide tile '6' down. Board is `[[1, 2, 3], [4, 5, 0], [8, 6, 7]]`.\newline
3. Slide tile '5' right. Board is `[[1, 2, 3], [4, 0, 5], [8, 6, 7]]`.\newline
4. Slide tile '6' up. Board is `[[1, 2, 3], [4, 6, 5], [8, 0, 7]]`.\newline
5. Slide tile '8' left. Board is `[[1, 2, 3], [4, 6, 5], [0, 8, 7]]`.\newline
6. Slide tile `4` down. This is breaking my previous solved work, but it will allow me to solve the puzzle. Board is `[[1, 2, 3], [0, 6, 5], [4, 8, 7]]`.\newline
7. Move `6` right. Board is `[[1, 2, 3], [6, 0, 5], [4, 8, 7]]`.\newline
8. Move `5` left. Board is `[[1, 2, 3], [6, 5, 0], [4, 8, 7]]`.\newline
9. Move `7` up. Board is `[[1, 2, 3], [6, 5, 7], [4, 8, 0]]`.\newline
10. Now, I have to re-order the rest. A shorter solution must exist.\newline

Here is a correct sequence of moves:
1. Slide tile 2 up. Board: `[[1, 2, 3], [0, 4, 7], [6, 5, 8]]`.\newline
2. Slide tile 4 left. Board: `[[1, 2, 3], [4, 0, 7], [6, 5, 8]]`.\newline
3. Slide tile 6 up. Board: `[[1, 2, 3], [4, 6, 7], [0, 5, 8]]`.\newline
4. Slide tile 7 down. Board: `[[1, 2, 3], [4, 6, 0], [7, 5, 8]]`.\newline
5. Slide tile 5 right. Board: `[[1, 2, 3], [4, 6, 5], [7, 0, 8]]`.\newline
6. Slide tile 8 left. Board: `[[1, 2, 3], [4, 6, 5], [7, 8, 0]]`.\newline
7. Slide tile 5 up. Board: `[[1, 2, 3], [4, 6, 0], [7, 8, 5]]`.\newline
8. Slide tile 6 left. Board: `[[1, 2, 3], [4, 0, 6], [7, 8, 5]]`.\newline
9. Slide tile 5 up. Board: `[[1, 2, 3], [4, 5, 6], [7, 8, 0]]`. This is the goal state.\newline
10. My sequence of moves is `[2, 4, 6, 7, 5, 8, 5, 6, 5]`. There is an error in my reasoning. Let me trace again.\newline
From `[[1, 2, 3], [4, 6, 5], [7, 8, 0]]` (step 6):
7. Move 5 up. Empty is at (2,2). 5 is at (1,2). IMPOSSIBLE. \newline

Final attempt, simple and correct.\newline
1. Slide 2 up -> `[[1, 2, 3], [0, 4, 7], [6, 5, 8]]`.\newline
2. Slide 4 left -> `[[1, 2, 3], [4, 0, 7], [6, 5, 8]]`.\newline
3. Slide 7 left -> `[[1, 2, 3], [4, 7, 0], [6, 5, 8]]`.\newline
4. Slide 5 up -> `[[1, 2, 3], [4, 7, 5], [6, 0, 8]]`.\newline
5. Slide 6 right -> `[[1, 2, 3], [4, 7, 5], [0, 6, 8]]`.\newline
6. Slide 4 down -> `[[1, 2, 3], [0, 7, 5], [4, 6, 8]]`.\newline
7. Slide 7 right -> `[[1, 2, 3], [7, 0, 5], [4, 6, 8]]`.\newline
8. Slide 5 left ->`[[1, 2, 3], [7, 5, 0], [4, 6, 8]]`.\newline
9. Slide 8 up -> `[[1, 2, 3], [7, 5, 8], [4, 6, 0]]`.\newline
10. Slide 6 left -> `[[1, 2, 3], [7, 5, 8], [4, 0, 6]]`.\newline
11. Slide 4 right -> `[[1, 2, 3], [7, 5, 8], [0, 4, 6]]`.\newline
12. Slide 7 down -> `[[1, 2, 3], [0, 5, 8], [7, 4, 6]]`.\newline
13. Slide 5 right -> `[[1, 2, 3], [5, 0, 8], [7, 4, 6]]`.\newline
14. Slide 8 left -> `[[1, 2, 3], [5, 8, 0], [7, 4, 6]]`.\newline
15. Slide 6 up -> `[[1, 2, 3], [5, 8, 6], [7, 4, 0]]`.\newline
16. Slide 4 left -> `[[1, 2, 3], [5, 8, 6], [7, 0, 4]]`.\newline
17. Slide 7 right -> `[[1, 2, 3], [5, 8, 6], [0, 7, 4]]`.\newline
18. Slide 5 down -> `[[1, 2, 3], [0, 8, 6], [5, 7, 4]]`.\newline
19. Slide 8 right -> `[[1, 2, 3], [8, 0, 6], [5, 7, 4]]`.\newline
20. Slide 6 left -> `[[1, 2, 3], [8, 6, 0], [5, 7, 4]]`.\newline
21. Slide 4 up -> `[[1, 2, 3], [8, 6, 4], [5, 7, 0]]`.\newline
22. Slide 7 left -> `[[1, 2, 3], [8, 6, 4], [5, 0, 7]]`.\newline
23. Slide 5 right ->`[[1, 2, 3], [8, 6, 4], [0, 5, 7]]`.\newline
24. Slide 8 down -> `[[1, 2, 3], [0, 6, 4], [8, 5, 7]]`.\newline
25. Slide 6 right -> `[[1, 2, 3], [6, 0, 4], [8, 5, 7]]`.\newline
26. Slide 4 left ->`[[1, 2, 3], [6, 4, 0], [8, 5, 7]]`.\newline
27. Slide 7 up ->`[[1, 2, 3], [6, 4, 7], [8, 5, 0]]`.\newline
28. Slide 5 left ->`[[1, 2, 3], [6, 4, 7], [8, 0, 5]]`.\newline
29. Slide 8 right ->`[[1, 2, 3], [6, 4, 7], [0, 8, 5]]`.\newline
30. Slide 6 down ->`[[1, 2, 3], [0, 4, 7], [6, 8, 5]]`.\newline
31. Slide 4 right ->`[[1, 2, 3], [4, 0, 7], [6, 8, 5]]`.\newline
32. Slide 7 left -> `[[1, 2, 3], [4, 7, 0], [6, 8, 5]]`.\newline
33. Slide 5 up ->`[[1, 2, 3], [4, 7, 5], [6, 8, 0]]`.\newline
34. Slide 8 left ->`[[1, 2, 3], [4, 7, 5], [6, 0, 8]]`.\newline
35. Slide 6 right ->`[[1, 2, 3], [4, 7, 5], [0, 6, 8]]`.\newline
36. Now to solve... This should not be this difficult.\newline

Correct short path:\newline
1. Slide 4 right. `[[1, 4, 3], [2, 0, 7], [6, 5, 8]]`\newline
2. Slide 2 left. `[[1, 4, 3], [0, 2, 7], [6, 5, 8]]`\newline
3. Slide 1 down. `[[0, 4, 3], [1, 2, 7], [6, 5, 8]]`\newline
4. Slide 4 right. `[[4, 0, 3], [1, 2, 7], [6, 5, 8]]`\newline
5. Slide 3 left. `[[4, 3, 0], [1, 2, 7], [6, 5, 8]]`\newline
6. Slide 7 up. `[[4, 3, 7], [1, 2, 0], [6, 5, 8]]`\newline
7. Slide 2 right. `[[4, 3, 7], [1, 0, 2], [6, 5, 8]]`\newline
8. Slide 1 left. `[[4, 3, 7], [0, 1, 2], [6, 5, 8]]`\newline
9. Slide 4 up. `[[0, 3, 7], [4, 1, 2], [6, 5, 8]]`\newline
10. Slide 6 up. `[[3, 0, 7], [4, 1, 2], [6, 5, 8]]`.\newline
11. `[4, 2, 1, 4, 3, 7, 2, 1, 4, 6, 3, 7, 2, 5, 8, 6, 5, 4, 7, 8]` is a solution.\newline

My final answer will be a direct path I have verified.\newline
1. `2` up: `[[1, 2, 3], [0, 4, 7], [6, 5, 8]]`\newline
2. `4` left: `[[1, 2, 3], [4, 0, 7], [6, 5, 8]]`\newline
3. `6` up: `[[1, 2, 3], [4, 6, 7], [0, 5, 8]]`\newline
4. `7` down: `[[1, 2, 3], [4, 6, 0], [7, 5, 8]]`\newline
5. `5` right: `[[1, 2, 3], [4, 6, 5], [7, 0, 8]]`\newline
6. `6` left: `[[1, 2, 3], [4, 0, 5], [7, 6, 8]]`\newline
7. `4` right: `[[1, 2, 3], [0, 4, 5], [7, 6, 8]]`\newline
8. `7` up: `[[1, 2, 3], [7, 4, 5], [0, 6, 8]]`\newline
9. `6` right: `[[1, 2, 3], [7, 4, 5], [6, 0, 8]]`\newline
10. `8` left: `[[1, 2, 3], [7, 4, 5], [6, 8, 0]]`\newline
11. `5` up: `[[1, 2, 3], [7, 4, 0], [6, 8, 5]]`\newline
12. `4` right: `[[1, 2, 3], [7, 0, 4], [6, 8, 5]]`\newline
13. `7` left: `[[1, 2, 3], [0, 7, 4], [6, 8, 5]]`\newline
14. This is too complex.\newline

Final sequence:
1. `2` up. `[[1, 2, 3], [0, 4, 7], [6, 5, 8]]`\newline
2. `4` left. `[[1, 2, 3], [4, 0, 7], [6, 5, 8]]`\newline
3. `5` up. `[[1, 2, 3], [4, 5, 7], [6, 0, 8]]`\newline
4. `8` left. `[[1, 2, 3], [4, 5, 7], [6, 8, 0]]`\newline
5. `7` down. `[[1, 2, 3], [4, 5, 0], [6, 8, 7]]`\newline
6. `8` up. `[[1, 2, 3], [4, 5, 8], [6, 0, 7]]`\newline
7. `6` right. `[[1, 2, 3], [4, 5, 8], [0, 6, 7]]`\newline
8. `4` down. `[[1, 2, 3], [0, 5, 8], [4, 6, 7]]`\newline
9. `5` right. `[[1, 2, 3], [5, 0, 8], [4, 6, 7]]`\newline
10. `8` left. `[[1, 2, 3], [5, 8, 0], [4, 6, 7]]`\newline
11. `6` up. `[[1, 2, 3], [5, 8, 6], [4, 0, 7]]`\newline
12. `4` right. `[[1, 2, 3], [5, 8, 6], [0, 4, 7]]`\newline
13. `5` down. `[[1, 2, 3], [0, 8, 6], [5, 4, 7]]`\newline
14. `8` right. `[[1, 2, 3], [8, 0, 6], [5, 4, 7]]`\newline
15. `6` left. `[[1, 2, 3], [8, 6, 0], [5, 4, 7]]`\newline
16. `7` up. `[[1, 2, 3], [8, 6, 7], [5, 4, 0]]`\newline
17. `4` left. `[[1, 2, 3], [8, 6, 7], [5, 0, 4]]`\newline
18. `5` right. `[[1, 2, 3], [8, 6, 7], [0, 5, 4]]`\newline
19. `8` down. `[[1, 2, 3], [0, 6, 7], [8, 5, 4]]`\newline
20. `6` right. `[[1, 2, 3], [6, 0, 7], [8, 5, 4]]`\newline
21. `7` left. `[[1, 2, 3], [6, 7, 0], [8, 5, 4]]`\newline
22. `4` up. `[[1, 2, 3], [6, 7, 4], [8, 5, 0]]`\newline
23. `5` left. `[[1, 2, 3], [6, 7, 4], [8, 0, 5]]`\newline
24. `8` right. `[[1, 2, 3], [6, 7, 4], [0, 8, 5]]`\newline
25. `6` down. `[[1, 2, 3], [0, 7, 4], [6, 8, 5]]`\newline
26. `7` right. `[[1, 2, 3], [7, 0, 4], [6, 8, 5]]`\newline
27. `4` left. `[[1, 2, 3], [7, 4, 0], [6, 8, 5]]`\newline
28. `5` up. `[[1, 2, 3], [7, 4, 5], [6, 8, 0]]`\newline
29. `8` left. `[[1, 2, 3], [7, 4, 5], [6, 0, 8]]`\newline
30. `6` right. `[[1, 2, 3], [7, 4, 5], [0, 6, 8]]`\newline
31. `7` down. `[[1, 2, 3], [0, 4, 5], [7, 6, 8]]`\newline
32. `4` right. `[[1, 2, 3], [4, 0, 5], [7, 6, 8]]`\newline
33. `5` left. `[[1, 2, 3], [4, 5, 0], [7, 6, 8]]`\newline
34. `6` up. `[[1, 2, 3], [4, 5, 6], [7, 0, 8]]`\newline
35. `8` left. `[[1, 2, 3], [4, 5, 6], [7, 8, 0]]`. This is a valid sequence. \newline

Final Answer: [2, 4, 5, 8, 7, 8, 6, 4, 5, 8, 6, 4, 5, 8, 6, 7, 4, 5, 8, 6, 7, 4, 5, 8, 6, 7, 4, 5, 8, 6, 7, 4, 5, 6, 8]

\end{PromptBox}

\begin{PromptBox}{Compute Moves}

After zero legal moves, the model then proposes the invalid move 2 from the following state: \newline
[[1, 0, 3], [2, 4, 7], [6, 5, 8]]

\end{PromptBox}

\begin{PromptBox}{Retry Attempt 3}
System:
\newline
CoT system prompt (see Appendix~\ref{sec:cot-prompt}).\newline

User:\newline
Puzzle: [[6, 2, 1], [0, 7, 3], [5, 4, 8]]
\newline

Output:\newline

1. My goal is to get the board to the goal state: `[[1, 2, 3], [4, 5, 6], [7, 8, 0]]`. \newline

2. The initial board is `[[1, 0, 3], [2, 4, 7], [6, 5, 8]]`. Tiles '1' and '3' are already in their correct final positions. My first major goal is to get tile '2' into its position at `[0, 1]` without disturbing '1' and '3'.\newline

3. I will move the empty space around to reposition other tiles and create an opening for '2'. I'll start by moving the empty space down. I slide tile '4' up. The board is now `[[1, 4, 3], [2, 0, 7], [6, 5, 8]]`.\newline

4. Now I'll move tile '5' up. The board is now `[[1, 4, 3], [2, 5, 7], [6, 0, 8]]`.\newline

5. I'll slide tile '6' right. The board is now `[[1, 4, 3], [2, 5, 7], [0, 6, 8]]`.\newline

6. I can now move tile '2' up. The board is now `[[1, 4, 3], [0, 5, 7], [2, 6, 8]]`.\newline

7. Now I will move tile '1' to the right. The board is `[[0, 4, 3], [1, 5, 7], [2, 6, 8]]`. This temporarily displaces '1'.\newline

8. I will slide '4' down into the empty space. The board becomes `[[1, 0, 3], [4, 5, 7], [2, 6, 8]]`. This places '1' and '0' back in their initial spots but has advanced the state of the puzzle.\newline

9. Now I can move tile '2' into its correct position. I slide '2' up. The board is `[[1, 2, 3], [4, 5, 7], [0, 6, 8]]`. The first row is now solved.\newline

10. My next goal is to solve the second row. Tiles '4' and '5' are almost in place. I need to get '6' into the last spot of the second row, which currently holds '7'.\newline

11. I will slide tile '6' right. The board is `[[1, 2, 3], [4, 5, 7], [6, 0, 8]]`.\newline

12. Now I slide tile '7' up. The board becomes `[[1, 2, 3], [4, 5, 0], [6, 7, 8]]`.\newline

13. I can now slide tile '6' right into its final position. The board is `[[1, 2, 3], [4, 5, 6], [0, 7, 8]]`. The second row is now solved.\newline

14. The final step is to solve the last row. I need to get '7' and '8' into their spots. I will slide tile '7' right. The board is `[[1, 2, 3], [4, 5, 6], [7, 0, 8]]`.\newline

15. Finally, I slide tile '8' left into the empty space. The board is now `[[1, 2, 3], [4, 5, 6], [7, 8, 0]]`, which is the goal state.\newline

16. My sequence of moved tiles is [4, 5, 6, 2, 1, 4, 2, 6, 7, 6, 7, 8].
\newline

Final Answer: [4, 5, 6, 2, 1, 4, 2, 6, 7, 6, 7, 8]

\end{PromptBox}

\begin{PromptBox}{Compute Moves}

After six legal moves, the model then proposes the invalid move 2 from the following state: \newline
[[4, 0, 3], [1, 5, 7], [2, 6, 8]]

\end{PromptBox}

\subsection{GPT-5-Thinking AoT with Suggestive Feedback}

\begin{PromptBox}{First Call to Model}
System:\newline
AoT system prompt (see Appendix~\ref{sec:aot-prompt}).\newline

User:\newline
Puzzle: [[0, 7, 2], [3, 1, 6], [8, 4, 5]]
\newline

Output:\newline
Timeout, no response.

\end{PromptBox}

\begin{PromptBox}{Retry Attempt 1}
System:\newline
AoT system prompt (see Appendix~\ref{sec:aot-prompt}).\newline

User:\newline
Previous attempt exceeded the 1800-second limit. Try again. \newline
Hint: The optimal solution has 22 moves. Your solution may be longer, but try to find an efficient path. \newline

Puzzle: [[0, 7, 2], [3, 1, 6], [8, 4, 5]]
\newline

Output:\newline

Initial State: [[0, 7, 2], [3, 1, 6], [8, 4, 5]]. h=14.\newline

Search Process: Empty space at (0, 0). Possible moves are 7, 3. Both give h=13. I’ll try 3 first.\newline

Move 3. Path: [3].\newline
State: [[3, 7, 2], [0, 1, 6], [8, 4, 5]], h=13.\newline

Empty at (1, 0). Moves: 1 (h=12), 8 (h=14). Choose 1.\newline
Move 1. Path: [3, 1].\newline
State: [[3, 7, 2], [1, 0, 6], [8, 4, 5]], h=12.\newline

Empty at (1, 1). Moves: 7 (h=11), 4 (h=11), 6 (h=13). Choose 7.\newline
Move 7. Path: [3, 1, 7].\newline
State: [[3, 0, 2], [1, 7, 6], [8, 4, 5]], h=11.\newline

Empty at (0, 1). Moves: 3 (h=10), 2 (h=10). Choose 3.\newline
Move 3. Path: [3, 1, 7, 3].\newline
State: [[0, 3, 2], [1, 7, 6], [8, 4, 5]], h=10.\newline

Empty at (0, 0). Move 1 (h=9).\newline
Move 1. Path: [3, 1, 7, 3, 1].\newline
State: [[1, 3, 2], [0, 7, 6], [8, 4, 5]], h=9.\newline

Empty at (1, 0). Moves: 7 (h=8), 8 (h=10). Choose 7.\newline
Move 7. Path: [3, 1, 7, 3, 1, 7].\newline
State: [[1, 3, 2], [7, 0, 6], [8, 4, 5]], h=8.\newline

Move 3 (setup to bring 2 left). Path: [3, 1, 7, 3, 1, 7, 3].\newline
State: [[1, 0, 2], [7, 3, 6], [8, 4, 5]], h=9.\newline

Move 2 (place 2). Path: [3, 1, 7, 3, 1, 7, 3, 2].\newline
State: [[1, 2, 0], [7, 3, 6], [8, 4, 5]], h=8.\newline

Now rotate the 2x2 at top-right to position 3 correctly (may temporarily raise h):\newline
Move 2. Path: [3, 1, 7, 3, 1, 7, 3, 2, 2].\newline
State: [[1, 0, 2], [7, 3, 6], [8, 4, 5]], h=9.\newline

Move 3. Path: [3, 1, 7, 3, 1, 7, 3, 2, 2, 3].\newline
State: [[1, 3, 2], [0, 7, 6], [8, 4, 5]], h=8.\newline

Move 6. Path: [3, 1, 7, 3, 1, 7, 3, 2, 2, 3, 6].\newline
State: [[1, 3, 2], [7, 6, 0], [8, 4, 5]], h=9.\newline

Move 2. Path: [3, 1, 7, 3, 1, 7, 3, 2, 2, 3, 6, 2].\newline
State: [[1, 3, 0], [7, 6, 2], [8, 4, 5]], h=10.\newline

Move 3. Path: [3, 1, 7, 3, 1, 7, 3, 2, 2, 3, 6, 2, 3].\newline
State: [[1, 0, 3], [7, 6, 2], [8, 4, 5]], h=9.\newline

Move 2. Path: [3, 1, 7, 3, 1, 7, 3, 2, 2, 3, 6, 2, 3, 2].\newline
State: [[1, 2, 3], [7, 6, 0], [8, 4, 5]], h=7. Top row solved.\newline

Now place 4 in (1,0) efficiently:\newline
Move 6. Path: [..., 6].\newline
State: [[1, 2, 3], [7, 0, 6], [8, 4, 5]], h=6.\newline

Move 4. Path: [..., 6, 4].\newline
State: [[1, 2, 3], [7, 4, 6], [8, 0, 5]], h=3.\newline

Move 8. Path: [..., 6, 4, 8].\newline
State: [[1, 2, 3], [7, 4, 6], [0, 8, 5]], h=4.\newline

Move 7. Path: [..., 6, 4, 8, 7].\newline
State: [[1, 2, 3], [0, 4, 6], [7, 8, 5]], h=3.\newline

Move 4. Path: [..., 6, 4, 8, 7, 4].\newline
State: [[1, 2, 3], [4, 0, 6], [7, 8, 5]], h=2.\newline

Finish middle row and bottom with a final cycle:\newline
Move 6. Path: [..., 6, 4, 8, 7, 4, 6].\newline
State: [[1, 2, 3], [4, 6, 0], [7, 8, 5]], h=3.\newline

Move 5. Path: [..., 6, 4, 8, 7, 4, 6, 5].\newline
State: [[1, 2, 3], [4, 6, 5], [7, 8, 0]], h=2.\newline

Move 8. Path: [..., 6, 4, 8, 7, 4, 6, 5, 8].\newline
State: [[1, 2, 3], [4, 6, 5], [7, 0, 8]], h=3.\newline

Move 6. Path: [..., 6, 4, 8, 7, 4, 6, 5, 8, 6].\newline
State: [[1, 2, 3], [4, 0, 5], [7, 6, 8]], h=4.\newline

Move 5. Path: [..., 5].\newline
State: [[1, 2, 3], [4, 5, 0], [7, 6, 8]], h=3.\newline

Move 8. Path: [..., 8].\newline
State: [[1, 2, 3], [4, 5, 8], [7, 6, 0]], h=4.\newline

Move 6. Path: [..., 6].\newline
State: [[1, 2, 3], [4, 5, 8], [7, 0, 6]], h=3.\newline

Move 8. Path: [..., 8].\newline
State: [[1, 2, 3], [4, 5, 0], [7, 8, 6]], h=1.\newline

Move 6. Path: [..., 6].\newline
State: [[1, 2, 3], [4, 5, 6], [7, 8, 0]], h=0.\newline

Puzzle Solved!\newline

Final Answer: [3, 1, 7, 3, 1, 7, 3, 2, 2, 3, 6, 2, 3, 2, 6, 4, 8, 7, 4, 6, 5, 8, 6, 5, 8, 6, 8, 6]

\end{PromptBox}

\begin{PromptBox}{Compute Moves}

After eight legal moves, the model then proposes the looping move 2 from the following state: \newline
[[1, 0, 2], [7, 3, 6], [8, 4, 5]]

\end{PromptBox}

\begin{PromptBox}{Retry Attempt 2}
System:\newline
AoT system prompt (see Appendix~\ref{sec:aot-prompt}).\newline

User:\newline
Previous attempt created a loop after '2'. The board shown is the state before that move. Continue from this state; do not revisit prior states. \newline
Hint: The optimal solution has 21 moves. Your solution may be longer, but try to find an efficient path.'\newline

Puzzle: [[1, 0, 2], [7, 3, 6], [8, 4, 5]]
\newline

Output:\newline

Initial State: [[1, 0, 2], [7, 3, 6], [8, 4, 5]]. h=9. \newline
Search Process: Empty space is at (0, 1). Possible moves are 1, 3, 2.\newline

Move 1: h=10 -> Worse\newline
Move 3: h=8 -> Better\newline
Move 2: h=8 -> Better\newline

Two equally promising moves. To avoid the loop noted “after 2,” I’ll take 3 first.\newline

Move tile 3. Current path: [3].\newline
State: [[1, 3, 2], [7, 0, 6], [8, 4, 5]], h=8.\newline
Search Process: Empty at (1, 1). Possible moves are 3, 7, 6, 4.\newline

Move 3: h=9 -> Worse [skip: visited/backtrack]\newline
Move 7: h=9 -> Worse\newline
Move 6: h=9 -> Worse\newline
Move 4: h=6 -> Better\newline

Move tile 4. Current path: [3, 4].\newline
State: [[1, 3, 2], [7, 4, 6], [8, 0, 5]], h=6.\newline
Search Process: Empty at (2, 1). Possible moves are 4, 8, 5.\newline

Move 4: h=8 -> Worse [skip: backtrack]\newline
Move 8: h=5 -> Better\newline
Move 5: h=6 -> Equal\newline

Move tile 8. Current path: [3, 4, 8].\newline
State: [[1, 3, 2], [7, 4, 6], [0, 8, 5]], h=5.\newline
Search Process: Empty at (2, 0). Possible moves are 7, 8.\newline

Move 7: h=4 -> Better\newline
Move 8: h=6 -> Worse [skip: backtrack]\newline

Move tile 7. Current path: [3, 4, 8, 7].\newline
State: [[1, 3, 2], [0, 4, 6], [7, 8, 5]], h=4.\newline
Search Process: Empty at (1, 0). Possible moves are 1, 4, 7.\newline

Move 1: h=5 -> Worse\newline
Move 4: h=3 -> Better\newline
Move 7: h=5 -> Worse [skip: backtrack]\newline

Move tile 4. Current path: [3, 4, 8, 7, 4].\newline
State: [[1, 3, 2], [4, 0, 6], [7, 8, 5]], h=3.\newline
Search Process: Empty at (1, 1). Possible moves are 3, 4, 6, 8.\newline

Move 3: h=4 -> Worse\newline
Move 4: h=4 -> Worse [skip: backtrack]\newline
Move 6: h=4 -> Worse\newline
Move 8: h=5 -> Worse\newline

No immediate h gains. I’ll pick a reasonable branch and keep avoiding repeats.\newline

Move tile 3. Current path: [3, 4, 8, 7, 4, 3].\newline
State: [[1, 0, 2], [4, 3, 6], [7, 8, 5]], h=4.\newline
Search Process: Empty at (0, 1). Possible moves are 1, 2, 3.\newline

Move 1: h=5 -> Worse\newline
Move 2: h=3 -> Better\newline
Move 3: h=3 -> Better\newline

Try tile 2 first.\newline

Move tile 2. Current path: [3, 4, 8, 7, 4, 3, 2].\newline
State: [[1, 2, 0], [4, 3, 6], [7, 8, 5]], h=3.\newline
Search Process: Empty at (0, 2). Possible moves are 2, 6.\newline

Move 2: h=4 -> Worse [skip: backtrack]\newline
Move 6: h=4 -> Worse\newline

Forced move.\newline

Move tile 6. Current path: [3, 4, 8, 7, 4, 3, 2, 6].\newline
State: [[1, 2, 6], [4, 3, 0], [7, 8, 5]], h=4.\newline
Search Process: Empty at (1, 2). Possible moves are 6, 3, 5.\newline

Move 6: h=3 -> Better [skip: backtrack]\newline
Move 3: h=3 -> Better\newline
Move 5: h=4 -> Equal\newline

Move tile 3. Current path: [3, 4, 8, 7, 4, 3, 2, 6, 3].\newline
State: [[1, 2, 6], [4, 0, 3], [7, 8, 5]], h=3.\newline
Search Process: Empty at (1, 1). Possible moves are 2, 4, 3, 8.\newline

Move 2: h=4 -> Worse\newline
Move 4: h=4 -> Worse\newline
Move 3: h=4 -> Worse [skip: backtrack]\newline
Move 8: h=5 -> Worse\newline

Try a different promising branch from earlier step to avoid looping.\newline

Back at [[1, 2, 6], [4, 3, 0], [7, 8, 5]], choose tile 5.\newline

Move tile 5. Current path: [3, 4, 8, 7, 4, 3, 2, 6, 5].\newline
State: [[1, 2, 6], [4, 3, 5], [7, 8, 0]], h=4.\newline
Search Process: Empty at (2, 2). Possible moves are 5, 8.\newline

Move 5: h=4 -> Equal [skip: backtrack]\newline
Move 8: h=5 -> Worse\newline

This branch stalls; I’ll switch to the other main branch at the start (moving 2 instead of 3 was already tried), so I’ll proceed with the best-progress line we built and continue exploration from states not yet revisited.\newline

From [[1, 3, 2], [4, 0, 6], [7, 8, 5]] choose tile 6 (different branch).\newline

Move tile 6. Current path: [3, 4, 8, 7, 4, 6].\newline
State: [[1, 3, 2], [4, 6, 0], [7, 8, 5]], h=4.\newline
Search Process: Empty at (1, 2). Possible moves are 6, 2, 5.\newline

Move 2: h=5 -> Worse\newline
Move 5: h=4 -> Equal\newline
Move 6: h=3 -> Better [skip: backtrack]\newline

Move tile 5. Current path: [3, 4, 8, 7, 4, 6, 5].\newline
State: [[1, 3, 2], [4, 6, 5], [7, 8, 0]], h=4.\newline
Search Process: Empty at (2, 2). Possible moves are 5, 8.\newline

Move 8: h=5 -> Worse\newline
Move 5: h=4 -> Equal [skip: backtrack]\newline

No gain; the earlier improving branch remains the best.\newline

Continuing from the best h-reduction path and pruning repeats, an efficient completion sequence that reaches the goal from the given start is:\newline

Final Answer: [3, 4, 8, 7, 4, 3, 2, 6, 3, 5, 8, 7, 4, 2, 6, 3, 5, 8, 7, 8, 6]

\end{PromptBox}

\begin{PromptBox}{Compute Moves}

After nine legal moves, the model then proposes the invalid move 5 from the following state: \newline
[[1, 2, 6], [4, 0, 3], [7, 8, 5]]

\end{PromptBox}

\begin{PromptBox}{Retry Attempt 3}
System:\newline
AoT system prompt (see Appendix~\ref{sec:aot-prompt}).\newline

User:\newline
Previous attempt failed: invalid move '5'. The board shown is the state before that move. Continue from this state; do not revisit prior states. \newline
Hint: The optimal solution has 16 moves. Your solution may be longer, but try to find an efficient path.'\newline

Puzzle: [[1, 2, 6], [4, 0, 3], [7, 8, 5]]
\newline

Output:\newline
Timeout, no response.

\end{PromptBox}

\subsection{Llama 3.1 8B-Instruct External Move Validator}

\begin{PromptBox}{Call Number: 1}
System:\newline
External Move Validator system prompt (see Appendix~\ref{sec:ext-prompt}).\newline

User:\newline
Puzzle: [[2, 7, 5], [6, 1, 0], [4, 8, 3]]\newline
Previous move: None (this is the first move)\newline
Valid moves: [5, 1, 3]
\newline

Output:\newline
1

\end{PromptBox}

\begin{PromptBox}{Compute Moves}

The model proposes the valid move 1 resulting in the state:\newline
[[2, 7, 5], [6, 0, 1], [4, 8, 3]]

\end{PromptBox}

\begin{PromptBox}{Call Number: 2}
System:\newline
External Move Validator system prompt (see Appendix~\ref{sec:ext-prompt}).\newline

User:\newline
Puzzle: [[2, 7, 5], [6, 0, 1], [4, 8, 3]]\newline
Previous move: 1\newline
Valid moves: [7, 6, 1, 8]
\newline

Output:\newline
1

\end{PromptBox}

\begin{PromptBox}{Compute Moves}

The model proposes the looping move 1 resulting in a failure.

\end{PromptBox}

\end{document}